\RequirePackage{fix-cm}
\documentclass[twocolumn]{svjour3}          % twocolumn
\smartqed  % flush right qed marks, e.g. at end of proof
\usepackage{array}
\newcolumntype{P}[1]{>{\centering\arraybackslash}p{#1}}
\newcolumntype{M}[1]{>{\centering\arraybackslash}m{#1}}
\usepackage{color, colortbl}
\definecolor{Gray}{gray}{0.9}
\usepackage{graphicx}
\usepackage{times}
\usepackage{epsfig}
\usepackage{epstopdf}
\usepackage{amsmath}
\usepackage{amssymb}
\usepackage{subfigure}
\usepackage{caption}
\usepackage{tabularx}
\usepackage{bm} % define this before the line numbering.
\usepackage[table]{xcolor}
\usepackage[normalem]{ulem}
\usepackage{algorithm}
\usepackage[noend]{algpseudocode}
\usepackage[section]{placeins}
\usepackage[affil-it]{authblk} 
\usepackage{etoolbox}
\usepackage{lmodern}
\usepackage{wrapfig}
\pdfoutput=1
\makeatletter
\def\BState{\State\hskip-\ALG@thistlm}
\renewcommand\section{\@startsection{section}{1}{\z@}%
	{-18\p@ \@plus -2\p@ \@minus -2\p@}%
	{8\p@ \@plus 2\p@ \@minus 2\p@}%
	{\normalfont\large\bfseries\boldmath
		\rightskip=\z@ \@plus 8em\pretolerance=10000 }}
\renewcommand\subsection{\@startsection{subsection}{2}{\z@}%
	{-12\p@ \@plus -2\p@ \@minus -2\p@}%
	{6\p@ \@plus 2\p@ \@minus 2\p@}%
	{\normalfont\normalsize\bfseries\boldmath
		\rightskip=\z@ \@plus 8em\pretolerance=10000 }}
\renewcommand\subsubsection{\@startsection{subsubsection}{3}{\z@}%
	{-6\p@ \@plus -2\p@ \@minus -2\p@}%
	{-0.3em \@plus -0.22em \@minus -0.1em}%
	{\normalfont\normalsize\bfseries\boldmath
		\rightskip=\z@ \@plus 8em\pretolerance=10000 }}
\patchcmd{\@maketitle}{\LARGE \@title}{\fontsize{16}{19.2}\selectfont\@title}{}{}
\makeatother
\definecolor{myGreen}{HTML}{33FF00}
\definecolor{myRed}{HTML}{FF3030}
\definecolor{myGrey}{HTML}{AA5555}
\definecolor{myWhite}{HTML}{FFFFFF}
\definecolor{maroon}{cmyk}{0,0.87,0.68,0.32}
\definecolor{petr}{HTML}{5555FF}
\definecolor{josef}{HTML}{FF3030}
\journalname{IJCV}
\begin{document}

\title{LCEval: Learned Composite Metric for Caption Evaluation}

%\subtitle{...and learning compact descriptors}

\titlerunning{$LCEval$}        % if too long for running head

\author{Naeha Sharif\and
	Lyndon White\and
	Mohammed Bennamoun\and
	Wei Liu\and
	Syed Afaq Ali Shah
}

\authorrunning{Naeha, Lyndon, Bennamoun, Wei, Afaq} % if too long for running head

\institute{Naeha Sharif, Mohammed Bennamoun, Wei Liu \at
			  Dept. of Computer Science, \\
              The University of Western Australia \\
              Perth, WA \\
              	\email{{naeha.sharif@research.uwa.edu.au}}\\
              	 \email{{\{mohammed.bennamoun,wei.liu\}@uwa.edu.au}}         
				\and
				Lyndon White \at
				Invenia Labs, \\
				Cambridge, United Kingdom \\
				\email{lyndon.white@invenialabs.co.uk}
           \and
           Syed Afaq Ali Shah \at
              Discipline of Information Technology, Mathematics \& Statistics, \\
              	Murdoch University \\
              	Australia \\
              		\email{afaq.shah@murdoch.edu.au}
}
\vspace{-5mm}
\date{Received: 02 Nov 2018  / Accepted: 23 Jul 2019 }

\maketitle

\begin{abstract}
Automatic evaluation metrics hold a fundamental importance in the development and fine-grained analysis of captioning systems. While current evaluation metrics tend to achieve an acceptable correlation with human judgements at the system level, they fail to do so at the caption level. In this work, we propose a neural network-based learned metric to improve the caption-level caption evaluation. To get a deeper insight into the parameters which impact a learned metric's performance, this paper investigates the relationship between different linguistic features and the caption-level correlation of the learned metrics. We also compare metrics trained with different training examples to measure the variations in their evaluation. Moreover, we perform a robustness analysis, which highlights the sensitivity of learned and handcrafted metrics to various sentence perturbations. Our empirical analysis shows that our proposed metric not only outperforms the existing metrics in terms of caption-level correlation but it also shows a strong system-level correlation against human assessments. 

\keywords{image captioning; automatic evaluation metric; neural networks; learned metrics; correlation; accuracy; robustness}
\end{abstract}

\section{Introduction}
Describing the visual world in natural language is a challenging task, which requires the ability to understand the visual content and generate well-formed textual descriptions. For humans, a brief glance is sufficient to understand the semantic meaning of a scene in order to describe the vast amount of visual details and subtleties \cite{you2018image}. Reasonable advancement has been made towards mimicking this human trait, but it is yet far from being achieved satisfactorily \cite{van2017room}, \cite{hodosh2016focused}. With the rapid progress in image captioning research \cite{kulkarni2013babytalk}, \cite{karpathy2014deep}, \cite{vinyals2015show}, \cite{xu2015show}, \cite{ordonez2016large}, \cite{yao2016boosting}, \cite{lu2017knowing}, \cite{you2018image}, the need for reliable and efficient evaluation methods has become increasingly pressing. Effective evaluation methodologies are necessary to facilitate system development and to identify potential areas for further improvements.
\begin{figure*}[t]
	\centering
	\includegraphics[width=1\textwidth]{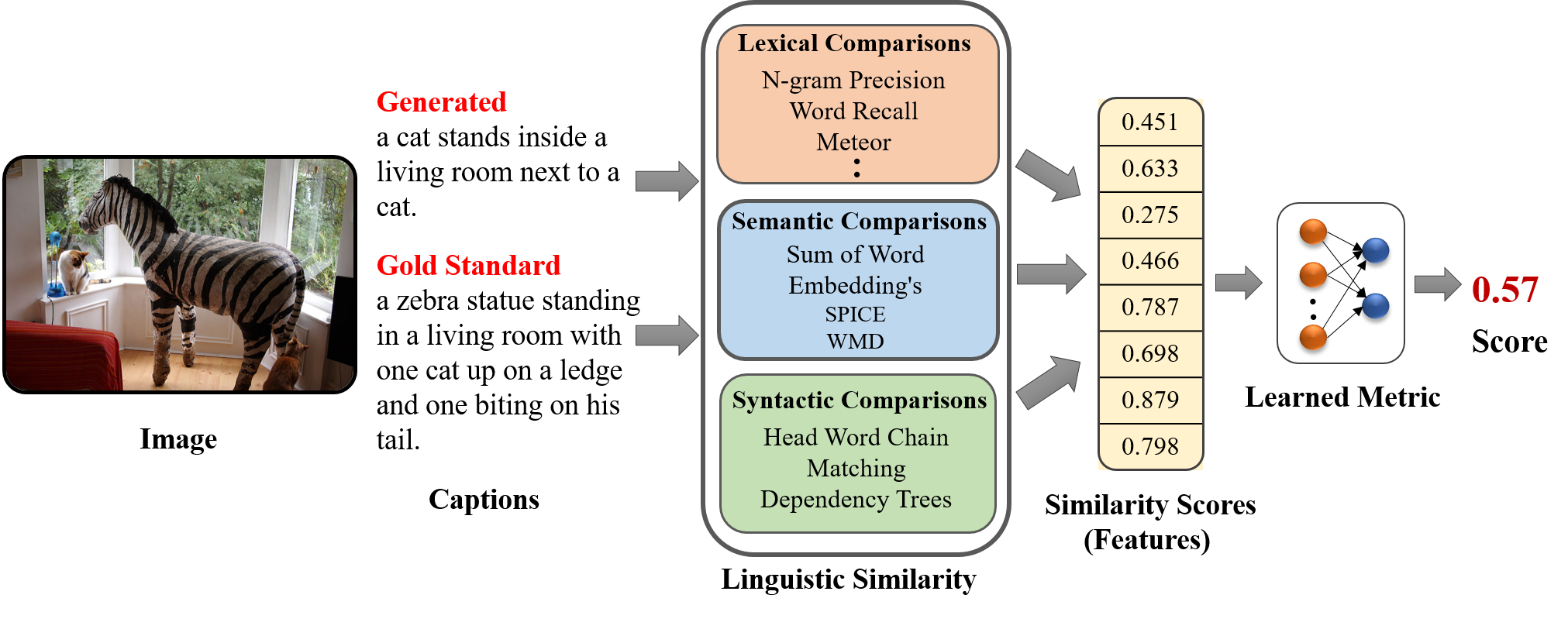}
	\vspace{-1.5em}
	\caption{An overview of the caption evaluation process, using a learned metric. Given a candidate caption and a set of reference captions, the learned metric evaluates the quality of the candidate caption based on a set of similarity scores (features). The features are extracted by carrying out various linguistic comparisons between the candidate and the references to assess the similarity at the lexical, semantic and syntactic levels. The more similar the generated caption to the reference, the better its quality.}
	\label{fig:PullUp}
	%\vspace{-1em}
	%\end{center}
\end{figure*}

Evaluating image descriptions is more complex than it is commonly perceived. This is mainly due to the diversity of acceptable solutions \cite{hodosh2013framing}. Human evaluation is often considered as the most reliable assessments for caption quality. It is however, resource intensive, subjective and hard to replicate. In contrast, automatic evaluation metrics are more efficient and can easily be replicated.
The downside of existing automatic  caption evaluation metrics is that they fail to reach the desired level of agreement with human judgements at the caption level \cite{elliott2014comparing}, \cite{kilickaya2016re}. Caption and system-level correlation statistics are two different types of evaluations which are carried out to examine the degree to which metric scores agree to the human judgements.  As obvious from names, caption-level correlation is measured on a caption-to-caption basis, whereas, system-level correlation is a coarse-grained evaluation on system-to-system basis. For computing system-level correlation, a metric score for each captioning system is computed, which is an aggregate of metric scores for all the captions generated by a system. Then, the correlation between the aggregate metric scores for the captioning systems and system-level human assessments is calculated.

While, system-level evaluation is more indicative of the relative qualities of different captioning systems, caption-level assessment reflects the difference in qualities of each caption generated by a captioning system and is more useful if we are interested in a confidence measure of the output captions. The problem of automatically evaluating captions is very complex, as many diverse solutions can be considered equally correct. Therefore, we need metrics that can evaluate the quality of various captions when one has to choose between a number of correct solutions. While most of the existing metrics such as METEOR \cite{banerjee2005meteor}, CIDEr \cite{vedantam2015cider} and SPICE \cite{anderson2016spice} show a moderate to strong correlation with human assessments at the system level \cite{anderson2016spice}, they fail to do so at the caption level. Moreover, the evaluation of some metrics wrongly show that the best machine models outperform humans in the image captioning task\footnote{http://cocodataset.org/\#captions-leaderboard} \cite{chen2015microsoft}, portraying an illusion that image captioning is close to being solved. \textit{This reflects the need to develop more reliable automatic metrics which capture the set of criteria that humans use in judging caption quality}.

Our motivation to form a composite metric is driven by the fact that the human judgement process involves assessments across various linguistic dimensions. We draw inspiration from the \textit{Machine Translation} (MT) literature, where learning paradigms have been proposed to create successful composite metrics \cite{bojar2016results}, \cite{bojar2017results}. A learning-based approach is useful because it offers a systematic way to combine various informative features. One possible approach to create a learned metric is to train a regression model. However, this requires a large corpus of manually scored captions, which is very expensive (in cost and time) to obtain. To mitigate this issue, we frame our learning problem as a classification task, in which the output is a simple binary variable indicating whether the caption is produced by a human or a machine. Due to the design of our classification task, the output is determined by the source of the candidate caption, needing no human assistance. On the other hand, if regression were attempted, manual assessments of each candidate caption would be required.

Our training criteria enables us to train our proposed network on a dataset of human and machine generated captions without the need of any human assessments of caption quality. Various publicly available datasets such as Flickr30k \cite{young2014image} and MSCOCO \cite{chen2015microsoft}, which have multiple human generated captions for each image can be re-purposed for training our learned metric. In other words, human written captions along with those generated by machine models can be used to train a network to distinguish between the two (human or machine). 

To the best of our knowledge, NNEval \cite{Sharif2018NNEval}, was the first neural-network based metric which was trained to distinguish between human and machine generated captions. In this article, we extend our previous work \cite{Sharif2018NNEval} by going deeper into issues related to learning-based metrics designed through classification. We particularly address concerns related to the sensitivity of the learned metrics to the features, the training data, the network architecture and the number of references. We investigate the impact of the various aforementioned parameters on caption-level correlation, accuracy and robustness of the proposed learned metric. 

In this work, we use an extended feature set, which includes syntactic features based on Head Word Chain Matching (HWCM), semantic feature based on Mean of Word Embeddings (MOWE) and simple lexical features such as n-gram precision, recall and Longest Common Subsequence (LCS), along with some other lexical and semantic features that are used in \cite{Sharif2018NNEval}.

Our main contributions in this work are:
\\\textbf{1.} The introduction of a new and extended set of features including semantic and syntactic features which have never been used before for image caption evaluation. 
\\\textbf{2.} A detailed analysis on the impact of using lexical, semantic, syntactic features, and their combinations on the caption-level correlation and classification accuracy of the learned metrics.
\\\textbf{3.} A comparison of performance variations on the learned metrics for different training datasets.
\\\textbf{4.} A detailed experimental analysis reflecting the various aspects of $LCEval$. Namely, its ability to correlate better with human judgements at the caption and the system level, its robustness to various distractions, and its performance by varying the number of reference captions.

\section{Related Work}
\subsection{Automatic Evaluation Metrics}
\begin{figure*}
	\centering
	\includegraphics[width=0.7\textwidth]{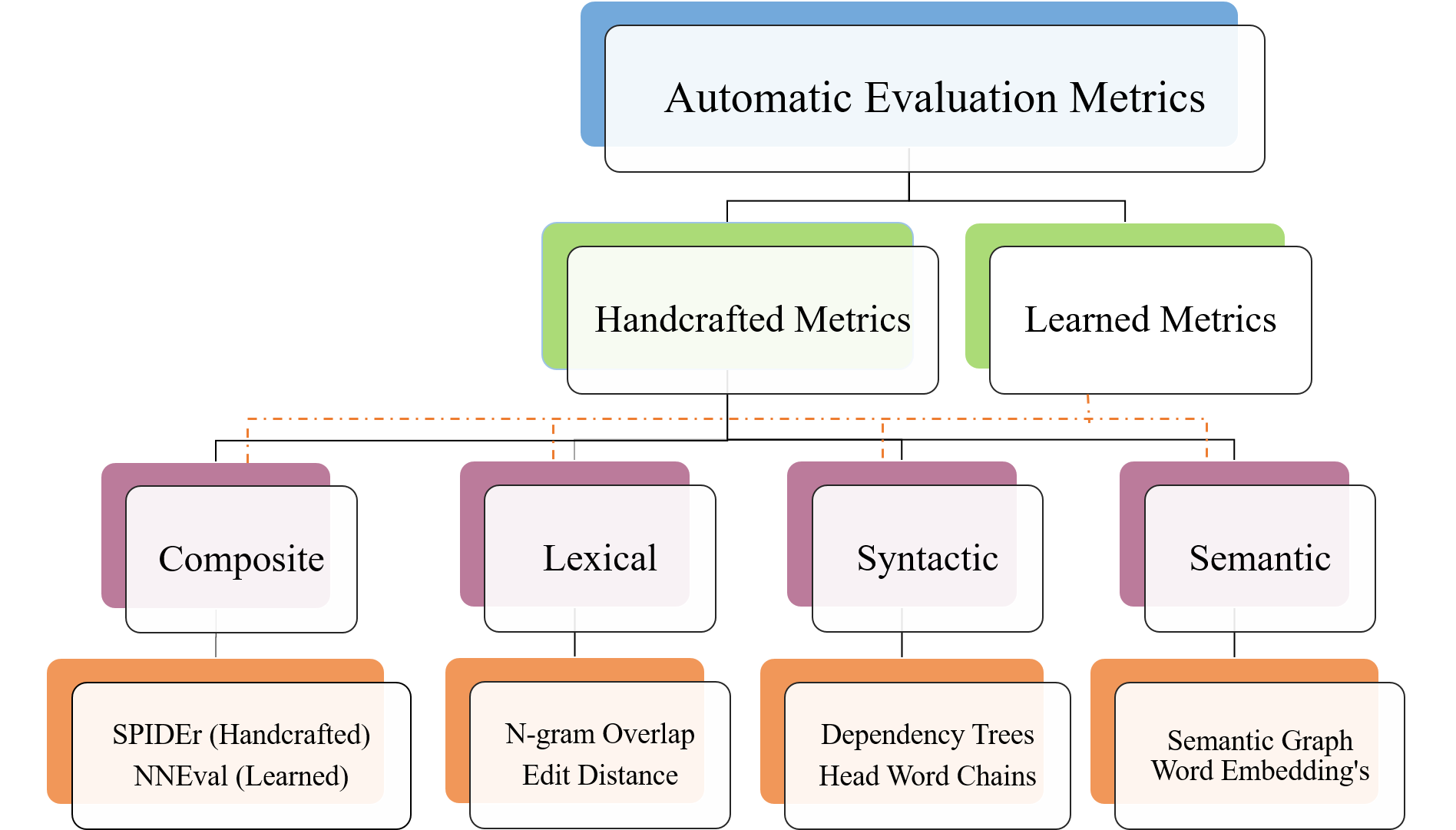}
	\vspace{-0.5em}
	\caption{Shows a broad taxonomy of evaluation metrics. The distinction between the different types of metrics or features is not a clear-cut, particularly for the lexical metrics, which often also overlap with semantic or syntactic metrics. }
	\label{fig:Taxonomy}
	\vspace{-1em}
	%\end{center}
\end{figure*}

The development of reliable automatic evaluation metrics is critical for the advancement of image captioning systems. While image captioning has drawn inspiration from the Machine Translation (MT) domain for encoder-decoder based captioning networks \cite{vinyals2015show}, \cite{xu2015show}, \cite{yao2016boosting}, \cite{you2016image}, \cite{lu2017knowing}, it has also benefited from the automatic metrics which were initially proposed to evaluate machine translations and text summaries \cite{papineni2002bleu}, \cite{banerjee2005meteor}, \cite{lin2004rouge}. More recently, various captioning-specific evaluation measures were developed \cite{vedantam2015cider}, \cite{anderson2016spice}, \cite{kilickaya2016re}, which have improved upon the correlation with human assessments. However, there is still a lot of room of improvement \cite{kilickaya2016re}, \cite{van2017room}. Figure~\ref{fig:Taxonomy} shows the broad taxonomy of caption evaluation metrics.

\subsection{Handcrafted Metrics}
Commonly used automatic metrics for image captioning, evaluate the caption quality by making deterministic measurements on the similarity between candidate and reference captions. These metrics tend to focus on specific aspects of correspondence, such as common sequences of words or the semantic likeness (e.g., using scene graphs as in SPICE \cite{anderson2016spice}). 

\textbf{Lexical measures} reward n-gram overlap between the candidate and reference sentences. These metrics can be categorized further into unigram-based methods for instance $BLEU_1$ and METEOR, and n-gram based methods such as $\text{ROUGE}_L$ and CIDEr. Some of these measures, apart from measuring the lexical similarity also take into account a certain degree of semantic and syntactic correspondence.  For example, METEOR which is predominantly a lexical measure, evaluates to some extent, the semantic aptness of the candidate through synonym matching and stemming. Similarly, n-gram measures such as $BLEU_4$ and CIDEr also take into account the syntactic similarity by capturing the word order. 

Since lexical measures rely primarily on the word overlap in order to capture caption quality, criticisms on their effectiveness is centered on the argument that \textit{n-gram overlap is neither necessary nor sufficient to measure the caption quality}. To overcome this limitation, other measures which take into consideration advanced linguistic features were proposed \cite{anderson2016spice}.

\textbf{Semantic measures} capture the semantic correspondence between sentences. SPICE is a semantic measure which uses scene graphs to measure the similarity between candidate and reference captions. SPICE has shown an improvement in the caption-level correlation compared to the commonly used measures (e.g., METEOR, CIDEr), however it tends to overlook the syntactic aptness of the candidates and is susceptible to be fooled by syntactically incorrect captions.  Recently Kilickaya et al. \cite{kilickaya2016re} showed that WMD can be a useful semantic metric for caption evaluation. WMD uses word embeddings to capture the semantic distance between documents. WMD also includes the lexical aptness through unigram matching.  
Purely \textbf{Syntactic measures}, which capture the grammatical similarity exist, such as Head Word Chain Matching \cite{liu2005syntactic} and have been used successfully in MT evaluation but have not been tested in the image captioning domain yet.

Existing handcrafted metrics fail to achieve adequate levels of correlation with human judgements at the caption-level. This reflects the fact that they do not fully capture the set of criteria that humans use to evaluate caption quality. One way to incorporate more features for evaluation is to combine various indicators, each of which focuses on a specific linguistic aspect, to form a fused metric (a.k.a. composite metric) \cite{liu2016improved}. This can be any combination of the metrics discussed above. For example, SPIDEr \cite{liu2016improved} is a linear combination of SPICE and CIDEr. It was found that optimising for SPIDEr scores resulted in higher quality captions, than optimizing for SPICE or CIDEr alone \cite{liu2016improved}. More generally, the metrics can come from any of the three categories discussed such as semantic, syntactic or lexical. Sharif et al. \cite{sharif2018learning} found in particular that covering all three categories improved the overall results.

\subsection{Learned Metrics}
Machine learning offers a systematic way to combine stand-alone handcrafted metrics or features into a unified measure. For learned metrics, the choice of features varies in the related literature, so does the learning algorithm. Combining meaningful linguistic features has shown promising results in metric evaluation campaigns, as reported in WMT (Workshop on Machine Translation) \cite{bojar2016results}, \cite{bojar2017results}. In \cite{Sharif2018NNEval}, we proposed the first neural network-based learned metric for caption evaluation, termed NNEval, combining the scores of different metrics through a learned framework. NNEval showed an improved performance over the commonly used stand-alone metrics and SPIDEr (composite metric) in terms of caption-level correlation. 

Recently, Cui et al. \cite{cui2018learning} also proposed a learned measure which blurs the line between caption evaluation and image captioning. In addition to the candidate caption, their system takes as input both a reference caption, as used in traditional caption evaluation; and the source image, as would be used as a context vector in image captioning systems. This is a promising approach as it provides the evaluation system with additional information that might be useful for evaluation. However, the incorporation of a sophisticated trained image processing system, does greatly increase the complexity of the overall evaluation. It thus, even more than usual, risks over-fitting to a particular domain. 

Furthermore, if there are particular image features such as color, shape and objects, that are readily detectable by the learned evaluator, then this risks to produce a biased judgement towards caption generation systems which pick-up on those same image features, even if they are of limited importance to the ideal caption. Cui et al. in \cite{cui2018learning} also did not report any results to analyze the extent to which the introduced image features help in improving the correlation with human judgements. Therefore, it is difficult to attribute their improved performance to the sentence embeddings or to the image features. An important contribution in our work is an attribution of quality to different features via ablation tests.

$LCEval$  scores the candidates based on their linguistic similarity with reference captions. Whereas in \cite{cui2018learning}, high-level learned features from a pre-trained LSTM \cite{hochreiter1997long} and CNN \cite{krizhevsky2012imagenet} are used for this purpose. Such high level features are not straightforward to interpret, which makes it harder to gain insight into the parameters used for evaluation. Moreover, high level image features are susceptible to pathological attacks (e.g., adversarial perturbations to images) and can lead to a poor performance of the automatic evaluator. Therefore, our proposed metric is based on the information which is accessible only to the evaluator, but not to the captioning systems. One of the most commonly used datasets for captioning, MSCOCO \cite{lin2014microsoft}, has a test set on which captioning systems can be evaluated. The images for this test set are publicly available but the corresponding reference captions which are used for the evaluation, are not available (they are hidden at this stage).

\section{$LCEval$: Proposed Learned Composite Metric}
\begin{figure*}
	\centering
	\includegraphics[width=1\textwidth]{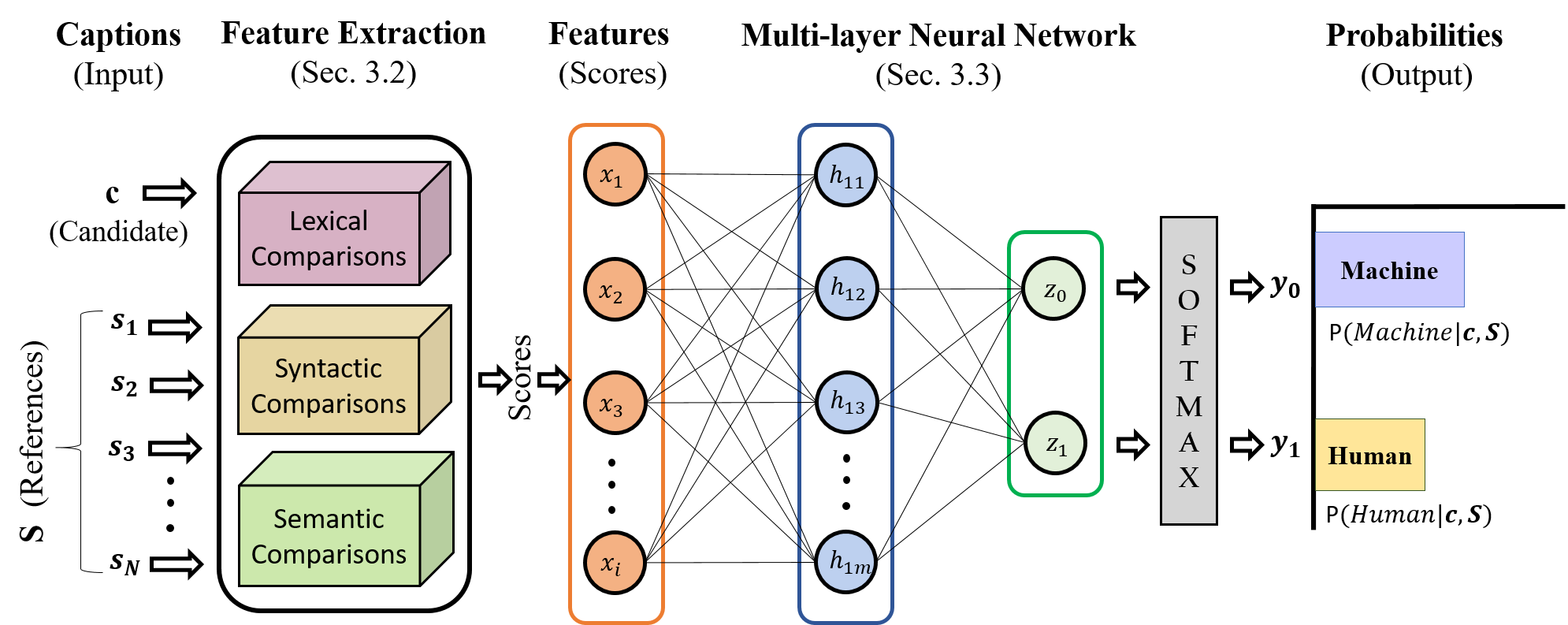}
	\vspace{-1.5em}
	\caption{Overall architecture of $LCEval$}
	\label{fig:Architecture}
	%\vspace{-1em}
	%\end{center}
\end{figure*}

In this Section, we describe the proposed metric in detail. The overall architecture of $LCEval$ is shown in Figure~\ref{fig:Architecture}. $LCEval$ is a generalised framework for learning composite caption-evaluation metrics. Our prior work on NNEval \cite{Sharif2018NNEval} is one particular instance of the $LCEval$ system. 

\subsection{Classification for Caption Evaluation}
To create a learning-based metric that is well aligned with human evaluations, we frame our learning problem as a classification task. We adopt a training criteria based on a simple question: ``is the candidate caption human or machine generated?'' In general, human generated captions are of superior quality and are easily distinguishable from the machine generated ones \cite{van2017room}, \cite{hodosh2016focused}. Image captioning would be a solved problem, if the outputs of image captioning systems were of such a high quality that they could not be distinguished from human generated captions. Exploiting this gap in quality, our trained classifier can set a boundary between human and machine produced captions. In order to obtain a continuous output score we use class-probabilities (Fig.~\ref{fig:Architecture}), instead of a class label. These probabilities represent the degree of confidence about a candidate belonging to one of these two classes i.e., human or machine. Thus, the resulting evaluator's output can be considered as some ``measure of believability'' that the input caption was human produced.

A more straightforward approach to create a learned metric would be to train a regression model.
However, this would require a large corpus of manually scored captions. The development of such a resource can be very difficult and time consuming \cite{kulesza2004learning}. Framing our learning problem as a classification task allows us to train our metric on any training set containing human written and machine generated captions for given images \cite{corston2001machine}. Various existing datasets \cite{plummer2015flickr30k}, \cite{chen2015microsoft} contain human written captions for their corresponding images and can be paired with various machine generated captions to serve as training examples. Thus, mitigating the need of obtaining expensive manual annotations. Moreover, such a dataset can be updated easily by including the outputs of more evolved captioning models without incurring any additional cost.

\subsection{$LCEval$ Features}
In our proposed framework, we feed a neural network with a set of numeric features representing lexical, semantic and syntactic similarities between the candidate \textbf{$c$} and a set of reference sentences \textbf{$S$}, as shown in Figure~\ref{fig:Architecture}. Each entity in the feature vector corresponds to a quality score generated by an individual measure which computes the similarity between the candidate and its corresponding references. While most of the measures that we use have a predefined way of handling multiple references per candidate, some others such as WMD, MOWE and HWCM compute their similarity using a single reference only. For such measures we evaluate the individual score against each reference and then use the maximum score (as done in \cite{denkowski2014meteor}). In our evaluations we group the features based on the linguistic properties that they capture as follows. 

\subsubsection{\textit{Lexical Comparisons against Reference \\Captions:}}
Measures for computing the lexical similarities between a candidate caption and its corresponding references are commonly used as stand-alone metrics in image captioning. Such as BLEU \cite{papineni2002bleu}, METEOR \cite{banerjee2005meteor}, ROUGE \cite{lin2004rouge} and CIDEr \cite{vedantam2015cider}. NNEval \cite{Sharif2018NNEval} used some of these measures as features to construct a learned metric. The lexical feature group (which is an extended version of the one used with NNEval) for this work consists of the following: 

\begin{itemize}
	\item\textit{N-gram Precision}: the proportion of n-grams in the candidate caption which match against the references, where (0${<}$n${<}$5)
	
	\item{\textit{Unigram Recall}}: the proportion of words (unigrams) in a reference caption which match the candidate caption. 
	
	\item{\textit{$\text{ROUGE}_L$}}: the longest common subsequence between the candidate and the reference captions.
	
	\item\textit{METEOR}: the unigram overlap between the candidate and the reference captions based on word meanings, exact forms and stemmed forms. 
	
	\item\textit{$\text{CIDEr}_D$}: the consensus between the candidate and the reference captions using n-gram matching. N-grams that are common in all the captions are down-weighted by computing the \textit{Term Frequency Inverse Document Frequency} (tf-idf) weighting. The mean cosine similarity between the n-grams of the reference and the candidate captions is referred to as the $CIDEr_n$ score. The final CIDEr score, which we use as a feature, is computed as the mean of $CIDEr_n$ scores, with $n=1, 2, 3, 4$.
\end{itemize}

\subsubsection{\textit{Semantic Comparisons against Reference Captions:}}
The semantic information is also crucial for the evaluation, since the captions using the same lexicon might not necessarily be similar to each other and vice versa. Therefore, we also use a group of features which capture the semantic relationship between the candidate and reference captions. These features are:

\begin{itemize}
	\item\textit{SPICE} estimates the caption quality by using ``scene graph'' representation for the candidate and the reference captions. The scene graph encodes the objects, attributes and relationships that are found in the captions. A set of logical tuples are formed by using possible combinations of the elements of the graphs to compute an F-score based on the conjunction of the candidate and reference caption tuples.
	
	\item\textit{WMD} measures the dissimilarity between two sentences as the minimum amount of distance that the embedded words of one sentence need to cover to reach the embedded words of the other sentence. More formally, each sentence is represented as a weighted point cloud of word embeddings \text{$ d \in R_N$}, whereas the distance between two words $i$ and $j$ is set as the Euclidean distance between their corresponding word2vec embeddings \cite{mikolov2013distributed}. To use it as a feature, we convert this distance score to a similarity by using a negative exponential. 
	
	\item\textit{MOWE (Mean of Word Embeddings)} captures the distance between the candidate and a reference caption by computing the cosine similarity between the sentence embeddings. Authors in \cite{White2015SentVecMeaning} found that the simple mean of word embeddings shows strong performances in capturing sentence meanings. The effectiveness of Mean of Word Embeddings (MOWE) when compared to other advanced models was as also noticed by \cite{Ritter2015Position}. We take the element-wise mean of the word embeddings over all the words in the caption to generate a vector which encodes the sentence meaning. We experimented with three different pre-trained word embeddings, and the details are shown in Table~\ref{table:embb}. Amongst the three embeddings, FastText \cite{bojanowski2017enriching} showed the most promising performance, as shown in Table~\ref{table:emm-results}. Thus, for all of our experiments, we used FastText. 
	 
\end{itemize}

\begin{table*}[h]
	\centering
	\caption{The details of pre-trained embeddings used in our experiments}
	\vspace{-1.0em}
	\label{table:embb}
	\begin{tabular}{|*{6}{c|}}
		\hline
		\textbf{Name} & \textbf{Source} &\textbf{Dimensions} &\textbf{Corpus} &\textbf{Corpus Size} &\textbf{Vocabulary Size} \\ \hline
		GloVE 840B 300d &	\cite{pennington2014glove} &	300	& Common Crawl &	8.40E+11&	2.20E+06\\
		\rowcolor{Gray}
		Word2vec Google 300d &	\cite{mikolov2013distributed}	& 300 &	Google News (100B) &	1.00E+11 &	3.00E+06\\
		FastText 300d &	\cite{bojanowski2017enriching}	& 300 &	Wikipedia &	3.96E+09 &	2.50E+06\\
		\hline
	\end{tabular}
	%\end{center}
\end{table*}

\begin{figure*}[t]
	\centering
	\includegraphics[width=1\textwidth]{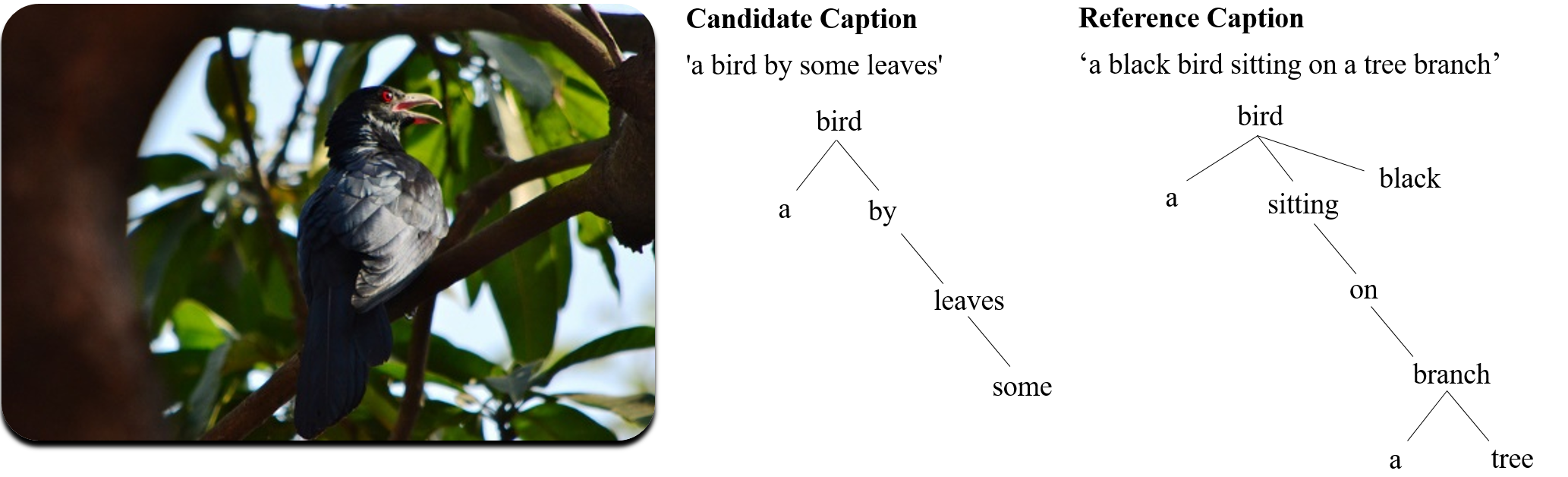}
	\vspace{-1.5em}
	\caption{Example of the dependency trees of the candidate and reference captions for an image sourced from MSCOCO dataset.}
	\label{fig:syntax}
	%\vspace{-1em}
	%\end{center}
\end{figure*}
\subsubsection{\textit{Syntactic Comparisons against Reference Captions:}}
We also match the syntactic constructs of the candidate and reference captions to assess the fluency of the candidate caption. This group consists of features based on Head Word Chain Matching (HWCM) proposed by Liu et al. in \cite{liu2005syntactic}, which have shown success in the case of MT evaluation.

\begin{itemize}
	\item\textit{Head Word Chain Matches (HWCM)}: these features capture the syntactic similarity between captions using the dependency tree structure of the sentences. Since one cannot expect to find the whole dependency tree of a candidate in the references, a sequence of head-modifier relationships is matched instead. An example of dependency trees of a reference and candidate caption is shown in Figure~\ref{fig:syntax}. The 2-word headword chains for the candidate include \textit{a bird, bird by, by leaves and leaves some}. The overlap of headword chains of length \textit{$u$} between the candidate and a reference caption normalised by the number of possible head-word chains of length \textit{$u$} in the candidate caption is used as a feature, where 1${<}$\textit{$u$}${<}$4. 
\end{itemize}

\subsection{Network Architecture and Learning task}
Given a candidate caption $c$ and a list of references $S=\{ s_1,s_2,s_3...s_N \}$, the goal is to classify the candidate caption as human or machine generated. We model this task using a feed-forward neural network, whose input is a fixed length feature vector \bm{$x$} = $\{x_1,x_2,x_3,..,x_i\} $, which we extract using the candidate caption and the corresponding reference captions (Section 3.2), and its output is the class probability, given as:
\begin{align}
\bm{y_{k}} & =\frac{e^{z_k}}{e^{z_0} + e^{z_1}},  k \in \{0,1\} 
\end{align} 
where \text{$z_k$} represents un-normalized class scores ($z_0$ and $z_1$ correspond to the machine and human class respectively). Our architecture has one hidden layer and the overall transformations in our network (Figure~\ref{fig:Architecture}) can be written as:
\begin{align}
\bm{h_{1}} & = \varphi( \bm{W_1} \bm{x}+ \bm{b_1}) \\
%\end{align}
%\begin{align}
\bm{z_k} & =   \bm{W_2} \bm{h_{1}}+ \bm{b_2}
\end{align}
\bm{$W_l$} and \bm{$b_l$}, are the weights and the bias terms between the input, hidden and output layers respectively.
 Where,\\ \text{$\bm{W_l} \in R^{N_l\times M_l}$}, \text{$\bm{b_l}\in R^{M_l}$} given\\ \text{$l\in\{1,2\}$}. \text{$\varphi(\cdot): R\rightarrow R$} is the non-linear activation function, given as:
\begin{align}
\varphi(v)  & = max(v, 0)
\end{align}
We use \text{$P(k=1|\bm{x})$} as our metric score, which is the probability of an input candidate caption being human generated. It can be formulated as:
\begin{align}
P(k=1|\bm{x}) & = \frac{e^{z_1}}{e^{z_0} + e^{z_1}}
\end{align}
The cross entropy loss for the training data with parameters, \text{$\bm{\theta} = (\bm{W_1,W_2,b_1},$} \text{$\bm{b_2} )$} can be written as:
\begin{align}
J_{\bm{\theta}} & = -\frac{1}{p}\sum_{t=1}^{p}\log(\frac{e^{z^t_{\tilde{y}}}}{e^{z^t_0} + e^{z^t_1}}) + \beta L
\end{align}
In the above equation  \text{$z^t_{\tilde{y}}$} is the activation of the output layer node corresponding to the true class \text{$\tilde{y}$}, given the input \bm{$x^t$}. Where, \text{$\beta L$} is a regularization term, that is commonly used to reduce model over-fitting. For our network we use \textit{$L_2$} regularization \cite{Ng:2004:FSL:1015330.1015435}.

\section{Experimental Methodology}
To analyse the performance of our proposed metric in comparison to the existing captioning metrics, we conducted  five sets of experiments. The goals of the experiments are to investigate: 1) the impact of network architecture, 2) the role of the features, 3) the accuracy and robustness of learned metrics, 4) the impact of using different number of references, and 5) the variation in the performance of metrics which are trained using different training examples. Following are the details of each experiment.

\begin{enumerate}
	
	\item \textbf{Impact of the network architecture}: We first measure the impact of learning weights to construct composite metrics as opposed to using unlearned weighting schemes in forming composite metrics As a baseline we choose SPIDEr (a linear combination of SPICE and CIDEr) and  $\text{$LCEval$}_{linear}$ which is a linear combination of all semantic, syntactic and lexical features which we use in this work.
	
	 We also evaluate neural network-based $LCEval$ with one, two and three hidden layers. On all models, we evaluate the Kendall's $\tau$ correlation score on our test set. See Section 5.1 for details.
	
	\item \textbf{Feature analysis}: We perform feature analysis on three different subgroups (Lexical, Semantic and Syntactic) to understand their impact on the learned metric and their individual contribution in improving the caption-level correlation. This analysis can be helpful in building better and customised evaluation metrics. We also examine the relationship between the classification accuracy and the caption-level correlation of the metrics when trained using different features. See Section 5.2 for details.
	
	\item \textbf{Classification accuracy and robustness}: We analyse the impact of features on the learned metrics to differentiate between pairs of captions. Moreover, we also compare the robustness of handcrafted and learned metrics against various sentence perturbations. See Section 5.3 and 5.4 for details.
	
	\item \textbf{Impact of the number of reference captions}: We investigate the impact of varying the number of reference captions on our learned metric $LCEval$. As per literature \cite{vedantam2015cider}, the performance of various handcrafted metrics is affected by the number of reference captions. See Section 5.5 for details.
	
	\item \textbf{Impact of training examples}: We compare the $LCEval$ metrics which are trained on machine generated captions from different systems. This is to determine how the quality and the quantity of the training examples affects the performance. Our hypothesis is that higher quality machine generated captions (as from a newer captioning system) will result in $LCEval$to perform better. See Section 5.6 for details.

	\item \textbf{Timing analysis}: We analyse the computation time of $LCEval$, and the existing measures to assess the practicality of using learned measures against the handcrafted ones . See Section 5.7 for details.
	
	\item \textbf{System-level correlation}: We evaluate the system-level correlation of our proposed metric against the human judgements collected in the 2015 COCO Captioning Challenge \cite{capeval2015}. See Section 5.8 for details.
	
\end{enumerate}

\begin{figure*}[h]
	\centering
	\includegraphics[width=1\textwidth]{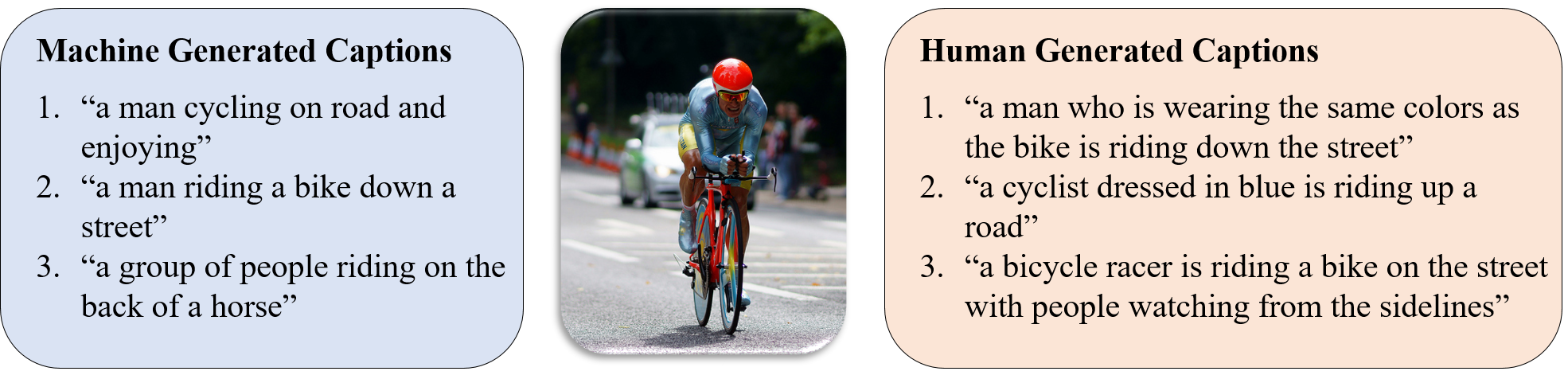}
	\vspace{-1.5em}
	\caption{shows an image from Flickr30k dataset with captions generated by machine models \cite{vinyals2015show}, \cite{xu2015show}, \cite{lu2017knowing} on the left and human generated captions for the corresponding image on the right.}
	\label{fig:Example of captions}
	%\vspace{-1em}
	%\end{center}
\end{figure*}

\subsection{Datasets for $LCEval$}
To train our metric, a dataset which contains both human and machine generated captions of each image is required.  Therefore, we create a training set by sourcing data from the Flickr30k dataset \cite{young2014image}. Flickr30k consists of 31,783 photos acquired from Flickr\footnote{https://www.flickr.com/}, each paired with 5 captions obtained through the Amazon Mechanical Turk (AMT). For each image in Flickr30k three out of five captions were chosen as human written candidate captions. The machine generated captions of the same images were obtained using three image captioning models, which achieved state-of-the-art performance at the time of their publication \cite{lu2017knowing},  \cite{vinyals2015show}, \cite{xu2015show}. 

\subsection{Dataset for Image Captioning Models}
The models that we use to obtain machine generated captions for our training set are: \textbf{1}) Show and Tell \cite{vinyals2015show}, \textbf{2}) Show, Attend and Tell (soft-attention) \cite{xu2015show}, and \textbf{3}) Adaptive Attention \cite{lu2017knowing}. We use publicly available official implementations for these captioning models and train them on the MSCOCO dataset \cite{chen2015microsoft}. MSCOCO dataset consists of training, validation and testing set with 82,783, 40,504 and 40,775 images, respectively. Each image in the training and validation sets is associated with five or more captions (collected through AMT), except for the testing set. We combine the MS COCO training and validation sets and use this combined set for the training of the captioning models, while reserving 10,000 image-caption pairs for validation and testing purposes. We train the image captioning models using the original experimental protocols to achieve close to their reported performances.

\subsection{Training Set for $LCEval$}
We use the three trained image captioning models discussed above (Sec. 4.2) to generate captions for the images of the Flickr30k dataset. For each image, we obtain three machine generated captions, one from each model. Moreover, we randomly choose three captions amongst the five human produced captions, which were originally paired with their respective image in Flickr30k, to use as human generated captions. This provides us with an equal number of human and machine generated candidate captions per image. Figure~\ref{fig:Example of captions} shows an example of human and machine produced candidate captions for a given image. In order to obtain reference captions for each candidate caption, we again utilize the human written descriptions of Flickr30k. For each machine-generated candidate caption, we randomly choose four out of five human written captions, which were originally associated with each image. Whereas, for each human-generated candidate caption, we select the remaining four out of five original AMT captions.

\begin{figure*}[h]
	\centering
	\includegraphics[width=1\textwidth]{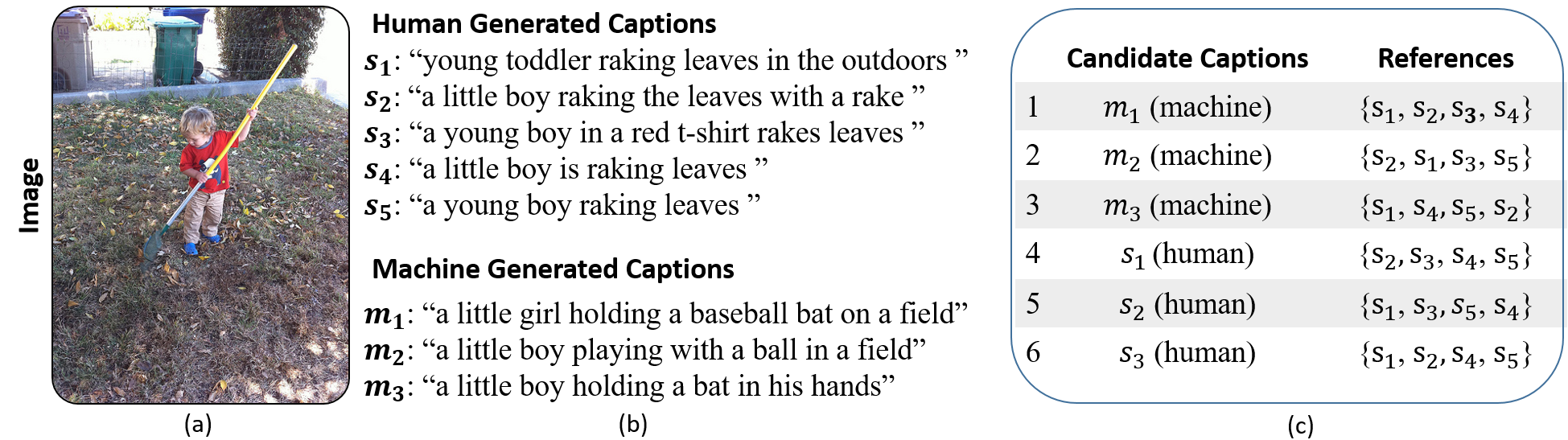}
	\vspace{-1.5em}
	\caption{shows an image (a), its corresponding human and machine generated captions (b), and candidate (human and machine generated captions) and reference pairing for the given image in the training set (c).}
	\label{fig:Candidate Reference Pairing}
	%\vspace{-1em}
	%\end{center}
\end{figure*}

In Figure~\ref{fig:Candidate Reference Pairing}, a possible pairing scenario is shown to demonstrate the distribution of the candidate and reference captions. If we select $s_1$ as the candidate human caption, we choose \{$s_2$, $s_3$, $s_4$, $s_5$\} as its corresponding reference captions. Whereas, when we select $m_1$ as a candidate machine caption, we randomly choose any of the four amongst \{$s_1$, $s_2$, $s_3$, $s_4$, $s_5$\} as references. While different sorts of pairing strategies can be explored, we leave that to future work. Moreover, the reason why we select four references for each caption is to exploit the optimal performance of each metric. Most of these metrics have been tested and reported to give a better performance with a larger number of reference captions \cite{papineni2002bleu}, \cite{denkowski2014meteor}, \cite{vedantam2015cider}, \cite{anderson2016spice}. 
\subsection{Validation Set for $LCEval$}
For our validation set, we draw data from Flickr8k \cite{young2014image} which consists of 8,092 images, each annotated with five human generated captions. The images in this dataset mainly focus on people and animals performing some action. This dataset also contains human judgements for a subset of 5,822 captions corresponding to 1000 images in total. Each caption was evaluated by three expert judges on a scale of 1 (the caption is unrelated to the image) to 4 (the caption describes the image without any errors). 

From our training set, we remove the captions of images which overlap with the captions in the validation and test sets (discussed in Sec. 4.5), leaving us with a total of 132,984 non-overlapping captions to train the proposed $LCEval$ model. We use this validation set for early stopping (see Sec. 4.6).

\subsection{Test Sets for $LCEval$}

\subsubsection{Correlation:} In order to measure the caption-level correlation of our proposed metric with human judgements, we use the COMPOSITE dataset \cite{aditya2017image} which contains human quality judgements for the  candidate captions and their corresponding image counterparts (Sec. 5.1). The images in this dataset are obtained from MS COCO, Flickr8k and Flickr30k datasets, whereas, their associated captions consist of human generated captions (sourced from the aforementioned datasets) and machine generated captions (using two captioning models \cite{aditya2017image}, \cite{karpathy2015deep}). The candidate captions of images are scored for correctness on the scale of 1 (low relevance) to 5 (high relevance) by AMT workers. From this dataset we use 7,996 single sentence based image captions, and evaluate the caption-level correlation performance of the commonly used handcrafted and proposed learned metric. 

We also use COMPOSITE dataset to evaluate the classification accuracy (to distinguish human and machine captions) of our learned metric (Sec. 5.2) and to study the impact of training examples (Sec. 5.6). For system-level correlation, we use human judgements collected in the 2015 COCO Captioning Challenge for 12 teams who participated in the 2015 COCO captioning challenge (Sec. 5.8).

\subsubsection{Pairwise Classification Accuracy:}
We measure the accuracy of metrics in terms of their ability to differentiate between pairs of candidate captions based on quality (see Sec. 5.3). A metric is considered accurate if it assigns higher scores to the captions preferred by humans in given pairs. For this experiment, we use PASCAL50S and ABSTRACT50S \cite{vedantam2015cider}. PASCAL50S contains human judgements for 4000 triplets of descriptions. Each triplet consists of one reference caption and two candidate captions.  The human judgements of PASCAL50S were collected through AMT, where the workers were asked to choose the candidate which is the most similar to the reference caption. Interestingly, the human judges were never shown the images corresponding to the references, therefore their judgements were based primarily on the sentence similarity between the candidate and the reference captions. 

Since the candidate sentences in PASCAL50S dataset consist of both human written and machine generated captions, the triplets are grouped into four categories i.e., Human-Human Correct (HHC), Human-Human Incorrect (HHI), Human-Machine (HM), Machine-Machine (MM). Each category comprises of 1000 triplets. The candidate captions in HHC and HHI categories are both human written, where in the former category both are correct and in the later, one is incorrect. In HM and MM categories, both candidate captions are correct. We also use PASCAL50S test set to investigate the impact of varying the number of reference captions (Sec. 5.5) and training examples (Sec. 5.6).

ABSTRACT50S contains 200 pairs of Human-Human Correct (HHC) captions and 200 pairs of Human-Human Incorrect (HHI) captions. In the HHC category, both captions are human written, correct and describe the same image. Whereas, in HHI category, while both captions are human written, one caption is correct and the other is incorrect.

\subsubsection{Robustness:} 
\begin{figure*}[t]
	\centering
	\includegraphics[width=1\textwidth]{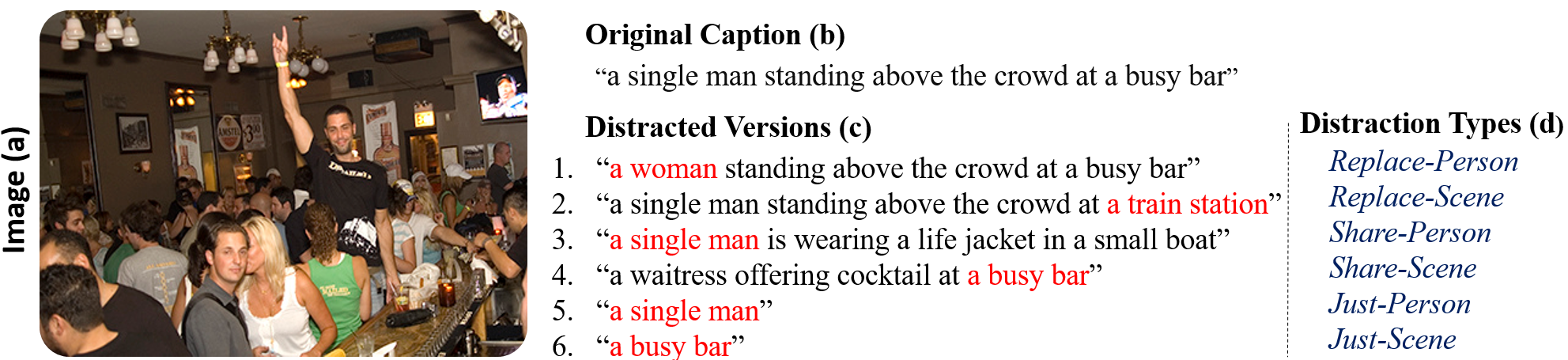}
	\vspace{-1.5em}
	\caption{shows an image (a), corresponding correct caption (b), distracted versions of the correct caption (c), and the type of distraction in each caption (d).}
	\label{fig:Distraction Example}
	%\vspace{-1em}
	%\end{center}
\end{figure*}

In \cite{hodosh2016focused}, authors introduced a dataset to perform a focused evaluation of image captioning systems with a series of binary forced-choice tasks. The data for each task contains images paired with one correct and one incorrect caption (distracted version of the correct caption). A robust image captioning metric should choose the correct caption over the perturbed one, to show that it can capture semantically significant changes in words and can identify when a complete sentence description is better than a single \emph{noun phrase}. In \cite{kilickaya2016re}, Kilickaya et al. used this dataset to perform a robustness analysis of various image captioning metrics. We use this same dataset, and report the performance on two more tasks, in addition to the four reported in \cite{kilickaya2016re}. Namely, \textbf{1}) \texttt{Replace Person} (RP), \textbf{2}) \texttt{Replace Scene} (RS), \textbf{3}) \texttt{Share Person} (SP), \textbf{4}) \texttt{Share Scene} (SS), \textbf{5}) \texttt{Just Person} (JP) and \textbf{6}) \texttt{Just Scene} (JS). An example of each of the six tasks is shown in Figure~\ref{fig:Distraction Example}. 

For the RS and RP task, given a correct caption for an image, the incorrect sentences (distractors) were constructed by replacing the scene/person (first person) in the correct caption with different scene/people. For the SP and SS tasks, the distractors share the same scene/task with the correct caption. However, the remaining part of the sentence is different. The JS and JP distractors only consist of the scene/person of the correct caption. We evaluate the metric scores for each correct and distracted version against the remaining correct captions that are available for each image in the dataset (Sec. 5.4).

\subsection{Training Parameters}

We use Adam optimizer  \cite{kingma2014adam} to train our network, with an initial learning rate of 0.0005 and a mini-batch size of 75.  We initialize the weights of our network by sampling values from a random uniform distribution \cite{glorot2010understanding}. $LCEval$ is optimized over the training set (Sec. 4.3) for a maximum of 800 epochs, and tested for classification accuracy on the validation set after each epoch.

\begin{figure}[h]
	\centering
	\includegraphics[width=80mm]{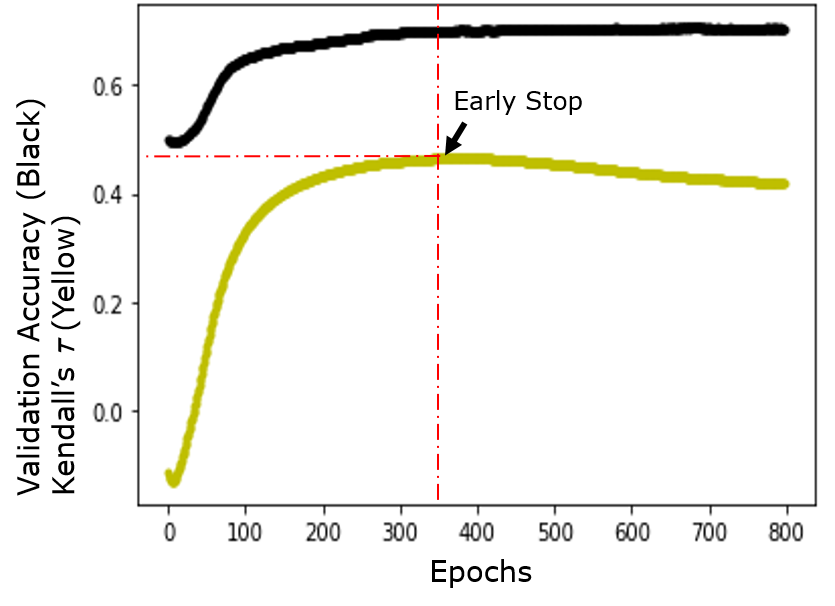}
	\vspace{-1.5em}
	\caption{Validation accuracy and model selection based on Kendall's {$\tau$}.}
	\label{fig:Early Stopping B}
	\vspace{-2em}
	%\end{center}
\end{figure}

While the loss function is used during the training period to maximize the classification accuracy, we are primarily interested in maximizing the correlation with human judgements. Therefore, we perform the early stopping based on Kendall's \textit{$\tau$}, which is evaluated on the validation set (Sec. 4.4) after each epoch. We thus terminate (early-stop) the training when the correlation is maximized as shown in Figure~\ref{fig:Early Stopping B}. Since each caption in the validation set is paired with three judgements, we use the most common value of these three judgements to evaluate the correlation coefficient.
The $LCEval$ architecture is implemented using TensorFlow library \cite{abadi2016tensorflow}.

\section{Results and Discussion}

In this section we discuss the results of various experiments that we performed to analyse the performance of our proposed metric. Our experiments allowed us to gain useful insights into various factors that can impact on the performance of the proposed learned metric for an improved caption evaluation.

\begin{table}[t]
	\centering
	\caption{A comparison of Kendall's {$\tau$} of metrics with alternative architectures}
	\vspace{-1em}
	\label{table:architectures}
	\begin{tabular}{  |M{1.15cm}||M{1.4cm}|M{1.15cm}|M{1.15cm}|M{1.12cm}| }
		\hline
		\textbf{Metrics} & \textbf{No Learning } & \textbf{1-layer NN}&  \textbf{2-layer NN}&  \textbf{3-layer NN} \\ \hline
		SPIDEr                 &        0.357         &       0.362        & \textbf{0.367}  & 0.365\\
		
		$LCEval$       &        0.273         &     0.375  & \textbf{0.418}        & 0.417\\
		\hline
	\end{tabular}
	%\end{center}
\end{table}

\subsection{The Choice of Network Architecture}

In this experiment, we compare the performance of the learned metrics with respect to the number of layers in their network. We also assess the effectiveness of combining metrics and/or feature scores into a composite metric through learning. For this purpose, we used three experimental settings: a linear combination of features (with no learning), one-layer neural network, and multilayer architectures which include the proposed two-layer (one hidden layer) architecture and a three-layer (two hidden layers)  architecture.  The learned and handcrafted composite metrics (baselines) are tested for their caption-level correlation using the COMPOSITE dataset (see Sec. 4.5) consisting of caption-level human judgements. We report the Kendall's {$\tau$} correlation coefficient, which is computed between the metric scores and human assessments scores for the test dataset.

The results are summarised in Table~\ref{table:architectures}. The first column shows the metrics, the second column shows the Kendall correlation of the metrics without any learning (using equal weights for all features and metric scores). The remaining columns show the Kendall's {$\tau$} of the learned metrics where the features are the same, but the number of layers in the network are different. The learned metrics perform noticeably better than the ones which involve no learning. The architecture based on the two-layers gives the optimal performance, adding or reducing a layer to it yields a drop of 0.01{$\tau$}  and 0.043{$\tau$} respectively in the case of {$LCEval$}. Therefore, we choose a two-layer architecture in our subsequent experiments.

\begin{table}[t]
	\centering
	\caption{Caption-level correlation of $LCEval$ on COMPOSITE test set using different pre-trained word embeddings for the semantic feature `MOWE'. For all the experiments we use FastText, as $\text{$LCEval$}_{FastText}$ achieves the best performance.}
	\vspace{-1.0em}
	\label{table:emm-results}
	\begin{tabular}{ |m{2.5cm}|*{3}{c|}}
		\hline
		\textbf{  Metric} & \textbf{Pearson} &\textbf{Spearman} &\textbf{Kendall} \\ \hline
		\textbf{ $\text{$LCEval$}_{GloVe}$}	& 	0.553	& 	0.530	&	0.414 \\
		\rowcolor{Gray}
		\textbf{ $\text{$LCEval$}_{Word2vec}$ }	& 	0.558	&	0.537	&	0.417 \\ 
		\textbf{ $\text{$LCEval$}_{FastText}$} &	\textbf{0.562}	&	\textbf{0.543}	&	\textbf{0.418} \\
		\hline
	\end{tabular}
	%\end{center}
\end{table} 

\subsection{Feature Analysis}
In Section 3.2 we gave an overview of the three categories of linguistic features that we use for our metric. In this experiment, we investigate the contribution of these features on the caption-level correlation of our proposed metric. First of all, we train the metric using all features and compare its performance to the commonly used handcrafted metrics on COMPOSITE dataset (Sec. 4.5.1). We report Pearson's $p$, Kendall's $\tau$ and Spearman's $r$ correlation coefficients for commonly used caption evaluation metrics along with a newer metric SPIDEr and our recently proposed learned metric NNEval \cite{liu2016improved}. The Table~\ref{table:correlation} shows that $LCEval$, trained using all the features, significantly outperforms other handcrafted metrics and our previously proposed metric NNEval, in terms of linear (Pearson) and rank based (Spearman and Kendall) correlations.

\begin{table}
	\centering
	\caption{Caption-level correlation of evaluation metrics with human quality judgments on the test set (Sec. 4.5.1). All p-values (not shown) are less than 0.001}
	\vspace{-1em}
	\label{table:correlation}
	\begin{tabular}{ |m{1.5cm}||M{1.5cm}|M{1.5cm}|M{1.5cm}|  }
		\hline
		\textbf{Metric}          & \textbf{Pearson} & \textbf{Spearman} & \textbf{Kendall} \\ \hline
		$\text{BLEU}_1$                   &      0.170       &       0.251       &      0.191       \\
		\rowcolor{Gray}
		$\text{BLEU}_4$ &      0.290       &       0.281       &      0.213       \\
		$\text{ROUGE}_L$              &      0.381       &       0.376       &      0.279       \\
		\rowcolor{Gray}
		METEOR &      0.411       &       0.447       &      0.341       \\
		$\text{CIDEr}_D$                    &      0.391       &       0.450       &      0.342       \\
		\rowcolor{Gray}
		SPICE  &      0.440       &       0.453       &      0.349       \\
		SPIDEr                   &      0.415       &       0.468       &      0.357       \\
		\rowcolor{Gray}
		WMD                  &      0.405       &       0.431       &      0.328       \\
		NNEval &      0.528      &       0.513       &      0.396       \\
		\rowcolor{Gray}
		\textbf{$LCEval$}        &  \textbf{0.562}  &  \textbf{0.543}   &  \textbf{0.418}  \\ \hline
	\end{tabular}
	%\end{center}
\end{table}

\begin{table}
	\centering
	\caption{A comparison of the impact of three different feature groups on the learned metric in terms of the Kendall's {$\tau$} correlation coefficient}
	\vspace{-1em}
	\label{table:correlationFeatureMetric}
	\begin{tabular}{ |m{2.5cm}||M{2cm}|M{2cm}|M{2cm}|  }
		\hline
		\textbf{Feature Group} & \textbf{Remove this group} & \textbf{Use only this group}   \\ \hline
		All Features           &             -              &            \textbf{0.418}               \\
		\rowcolor{Gray}
		Semantic               &           0.387            &             0.372               \\
		Lexical                &           0.366            &            0.409               \\
		\rowcolor{Gray}
		Syntactic              &           0.413            &            0.298               \\ \hline
	\end{tabular}
	%\end{center}
\end{table}
Next, we train our metric using the features belonging to each of the three main groups (Lexical, Semantic or Syntactic) separately  i.e., using one group of features at a time. Then, we train the metric by keeping the features from any two groups and leaving out the third one.  Table~\ref{table:correlationFeatureMetric} presents the variations in the Kendall's {$\tau$} correlation coefficients on COMPOSITE test set (Sec. 4.5.1), highlighting the significance of each category of features. Our choice of correlation coefficient (Kendall's {$\tau$}) for reporting caption-level correlation is consistent with \cite{cui2018learning} and \cite{anderson2016spice}. Out of the three groups of features, the learned metric relies heavily on the lexical comparisons against caption references. Using just the features of this group, the caption-level correlation stands at a significant 0.409. Whereas, without using lexical features, a steep drop in correlation is observed. Removing the features of the other groups also lowers the correlation, but the drop is marginal when we exclude syntactic features. However, this does not mean that syntactic features are not informative, since the highest correlation of 0.418 is only achieved by using all the features (including syntactic ones). See Appendix for qualitative results and the extended version of Table~\ref{table:correlationFeatureMetric}.  

\begin{table}
	\centering
	\caption{Kendall's {$\tau$} correlation of HWCM score for different head word chain lengths on the test set (Sec 4.5.1). Notice that it performs better than some of the metrics that are reported in Table~\ref{table:correlation}}
	\vspace{-1em}
	\label{table:Syb=ntacticFeaturesWL}
	\begin{tabular}{ |m{2cm}||M{0.9cm}|M{0.9cm}|M{0.9cm}|M{0.9cm}| }
		\hline
		Chain Depth  &1 &2 &3 &4 \\ \hline
		HWCM                &  0.292 & \textbf{0.297} & 0.293 & 0.291        \\ \hline
	\end{tabular}
	%\end{center}
\end{table}
Note that HWCM based syntactic features have never been used for caption evaluation. We report the Kendall's {$\tau$} of the scores generated by these measures against the human judgements on our test. The standalone HWCM features that use dependency relationships correlate better with human judgements as compared to the commonly used metrics, such as BLEU and $\text{ROUGE}_L$.

\begin{figure}[h]
	\centering
	\includegraphics[width=80mm]{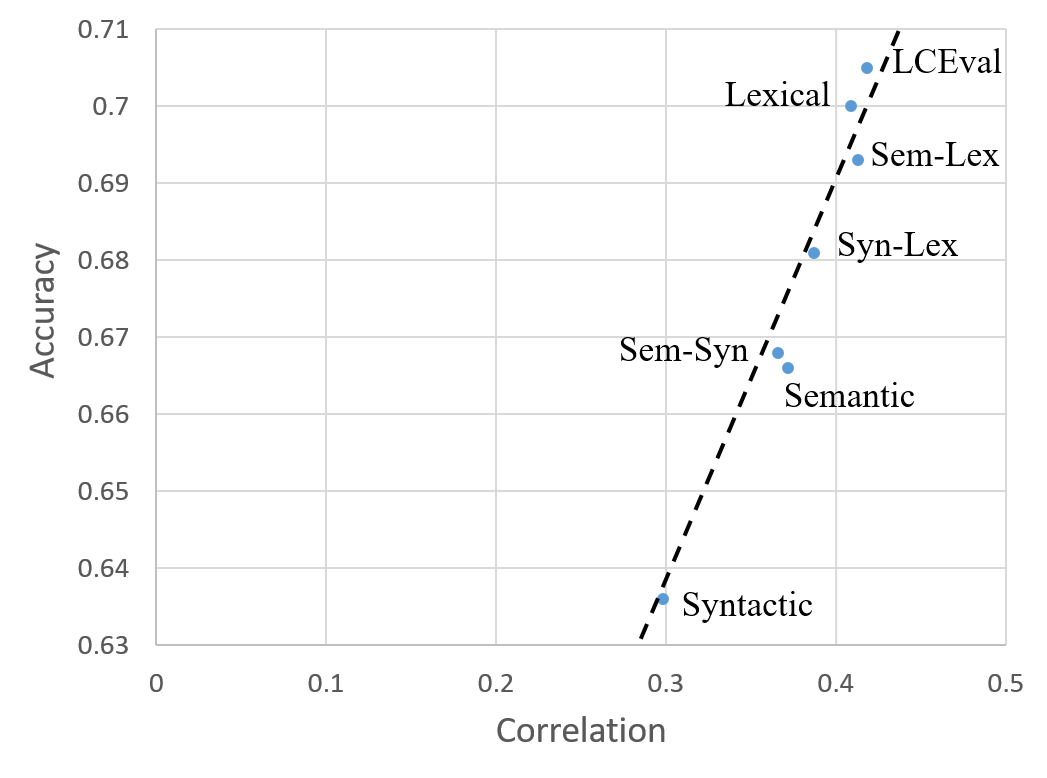}
	\vspace{-1.5em}
	\caption{Relationship between caption-level correlation (Kendall's {$\tau$}) and accuracy of caption evaluation metrics trained on different feature sets.}
	\label{fig:CorVsAccu}
	%\vspace{-1em}
	%\end{center}
\end{figure}

\subsubsection{Classification Accuracy \textit{Vs.} Caption-level Correlation}
We also analyse the relationship between the Kendall's {$\tau$} correlation coefficient and the classification accuracy of the metrics trained using different feature sets. We evaluate both, the caption-level correlation and the validation accuracy on COMPOSITE dataset (Sec. 4.5.1). As shown in Figure~\ref{fig:CorVsAccu}, metrics showing a higher correlation also achieve a higher classification accuracy. From the same figure, one can also observe a linear relationship between correlation and classification accuracy. $LCEval$ achieves the best overall correlation, as well as the best classification accuracy on the test set. This illustrates that a metric which achieves a higher classification accuracy also attains a stronger correlation with human judgements.

\subsection{Pairwise Classification Accuracy}

We follow the framework introduced in \cite{vedantam2015cider} to analyse the ability of metrics to discriminate between a pair of captions with reference to the ground truth captions. We follow the same original approach of \cite{vedantam2015cider}, using PASCAL50S (Sec. 4.5.2) to assess the accuracy of the metrics and report them in Table~\ref{table:pascal}. The slight variation from the previously reported results in \cite{vedantam2015cider} and \cite{anderson2016spice} might be due to the randomness in the choice of reference captions. 

The results in Table~\ref{table:pascal} show the accuracy of the handcrafted metrics as well as the learned metrics, which were trained using three different feature sets. The top segment of the table shows the results of the handcrafted metrics, the middle segment shows the results of the learned metrics, designed using specific set of features. The last segment shows the learned metric $LCEval$ trained using all the features. In the middle segment of Table~\ref{table:pascal}, the names indicate the group of features that are used to train that metric. For example \textit{Semantic} means that the metric is trained using only the semantic features. NNEval \cite{Sharif2018NNEval} was trained using a group of semantic and lexical features as discussed in \cite{Sharif2018NNEval}. Thus, NNEval can be regarded as a special case of the \textit{Sem-Lex} category. Though the lexical and semantic features used in NNEval are different from the ones used in this work.

It can be observed from Table~\ref{table:pascal} that on average, the learned metrics achieve a higher accuracy (80.6 and 80.5 in the case of \textit{Semantic} and NNEval respectively) compared to the handcrafted metrics. Though the learned metrics were trained using a simple training criteria i.e., to distinguish between good and bad quality captions, they still achieve reasonable accuracy in ranking captions which vary in quality. Another point to be noted is that the metrics that are trained with different features might learn different classification boundaries, which would impact on their ability to differentiate between caption-pairs. Moreover, there is no single metric, either learned or handcrafted, which produces the best performance across all categories. This shows that features have to be engineered for each task specifically if a high performance is required for that particular task. 
\begin{table}[t]
	\centering
	\caption{Comparative accuracy results on four kinds of caption pairs tested on PASCAL50S}
	\vspace{-1em}
	\label{table:pascal}
	\begin{tabular}{|m{1.5cm}|*{5}{c|}}		\hline
		\textbf{Metric}            & \textbf{HHC}   & \textbf{HHI}   & \textbf{HM}   & \textbf{MM}   & \textbf{AVG}  \\ \hline
		$\text{BLEU}_1$                     & 53.5          & 95.6          & 91.1          & 57.3          & 74.4          \\
		\rowcolor{Gray}
		$\text{BLEU}_4$   & 53.7          & 93.2          & 85.6          & 61.0          & 73.4          \\
		ROUGE                    & 56.5          & 95.3          & 93.4          & 58.5          & 75.9          \\
		\rowcolor{Gray}
		METEOR   & 61.1          & 97.6          & \textbf{94.6} & 62.0          & 78.8          \\
		$\text{CIDEr}_D$                      & 57.8          & 98.0          & 88.8          & 68.2          & 78.2          \\
		\rowcolor{Gray}
		SPICE    & 58.0          & 96.7          & 88.4          & 71.6          & 78.7          \\
		SPIDEr            & 56.7          & 99.0          & 91.0          & 69.1          & 78.8          \\
		\rowcolor{Gray}
		WMD            & 56.2          & 98.4          & 91.7          & 71.5          & 79.5         \\
		\hline\hline
		\rowcolor{Gray}
		Semantic & 59.7          & 98.7          & 91.6          & \textbf{72.4} & \textbf{80.6} \\
		Syntactic                  & 58.6          & 95.5          & 91.1          & 60.6          & 76.4          \\
		\rowcolor{Gray}
		Lexical  & 62.8          & 96.1          & 82.4          & 60.1          & 75.3          \\
		Sem-Syn                    & 57.1          & \textbf{99.2} & 91.2          & 68.4          & 79.0          \\
		\rowcolor{Gray}
		Sem-Lex  & 62.7          & 97.8          & 90.1          & 63.8          & 78.6          \\
		Syn-Lex                    & 63.4          & 97.5          & 92.4          & 62.2          & 78.9          \\ 
		\rowcolor{Gray}
		NNEval   & 60.4          & 99.0          & 92.1          & 70.4          & 80.5          \\ \hline \hline
		\textbf{$LCEval$}          & \textbf{64.0} & 97.6          & 90.1          & 65.4          & 79.3          \\ \hline
	\end{tabular}
	%\end{center}
\end{table}

The \textit{Semantic} learned metric achieves the best accuracy in differentiating between Machine-Machine Correct captions. This metric learned to differentiate between human and machine captions based on their semantic properties, and is able to distinguish between two machine generated captions by exploiting the very same (semantic) properties. It can also be observed that amongst the handcrafted metrics, SPICE, which itself is a semantic measure, is the best at classifying machine from machine captions. This portrays the importance of using semantic information for assessing machine captions. The machine captions for PASCAL50S were sourced from five different captioning systems i.e., Babytalk \cite{kulkarni2013babytalk}, Midge \cite{mitchell2012midge}, Story \cite{farhadi2010every}, Video \cite{rohrbach2013translating} and Video+ \cite{rohrbach2013translating}, all of which usually generate syntactically correct captions. The candidate captions differ in terms of adequacy (mentioning of the correct content). Since SPICE primarily focuses on the visual content (objects, attributes and relations) rather than syntax, it does a good job at differentiating between such captions.

HM is the only category in which handcrafted features perform better than the learned metrics. Surprisingly $\text{BLEU}_1$, which is a simple word matching based metric, performs better than other handcrafted measures, such as $\text{CIDEr}_D$ and SPICE. METEOR, which is an advanced version of unigram matching, uses synonyms and lemmas and achieves the best accuracy of 94.6\%.
\begin{table}[t]
	\centering
	\caption{Comparative accuracy results on two kinds of caption pairs tested on ABSTRACT50S}
	\vspace{-1.0em}
	\label{table:abstract505}
	\begin{tabular}{|m{1.5cm}|*{3}{c|}}
		\hline
		\textbf{Metric} & \textbf{HHC} & \textbf{HHI} & \textbf{AVG}\\ \hline
		$\text{BLEU}_1$		&	49.0 &	88.5 &	68.8 \\
		\rowcolor{Gray}
		$\text{BLEU}_4$		&	51.5 &	81.0 &	66.3 \\
		$\text{ROUGE}_L$	&	52.0 &	82.0 &	67.0 \\
		\rowcolor{Gray}
		METEOR		&	48.5 &	46.5 &	47.5 \\
		$\text{CIDEr}_D$		&	50.5 &	78.5 &	64.5 \\
		\rowcolor{Gray}
		SPICE		&	56.0 &	86.5 &	71.3 \\
		SPIDEr		&	58.0 &	91.5 &	74.8 \\
		\rowcolor{Gray}
		WMD		&	50.0 &	93.5 &	71.8 \\
		\hline\hline
		\rowcolor{Gray}
		Semantic	&	53.0 &	\textbf{94.0} &	73.5 \\
		Syntactic	&	\textbf{64.5} &	88.5 &	\textbf{76.5} \\
		\rowcolor{Gray}
		Lexical	&	55.5 &	86.5 &	71.0 \\
		Sem-Syn	&	52.5 &	93.5 &	73.0 \\
		\rowcolor{Gray}
		Sem-Lex	&	57.5 &	91.0 &	74.3 \\
		Syn-Lex	&	57.5 &	86.0 &	71.8 \\
		\rowcolor{Gray}
		NNEval		&	55.0 &	86.0 &	70.4 
		\\ \hline \hline
		$LCEval$		&	58.3 &	\textbf{94.0} &	76.1 \\
		\hline
	\end{tabular}
	%\end{center}
\end{table}
To our surprise $LCEval$ showed a comparatively low performance in HM category. Therefore, we did some further investigation to understand the reasons behind this issue. In the HM category, around 98 out of 1000 machine captions were misclassified by $LCEval$ to be better than the human generated captions. Since the classification between human and machine captions is not an easy task, even for humans, we investigated the human judgements in this category and found that for a total of 1000 pairs of human and machine captions, only 177 caption pairs have a unanimous human judgement (i.e., all the judges agreed on one caption being better than the others). However, for rest of the 823 pairs, at least one human judge disagreed with the other judges (i.e., in these cases the humans were fooled by the machine examples, or may be the human generated ones were not of that good quality to be easily judged better than the machine captions).   

In the HHI category, learned metrics trained with a \textit{Sem-Syn} group of features attain the highest accuracy of 99.2\%. In this category, all the metrics except for $\text{BLEU}_4$ achieve an accuracy greater than that of 95\%, showing that all metrics are good at picking up differences in such caption pairs. 

HHC is one of the toughest category, because it requires a metric to detect subtle differences in caption quality. It can be observed that four of the learned metrics (\textit{Lexical} (62.8\%), \textit{Sem-Lex} (62.7\%), \textit{Syn-Lex} (63.4\%) and $LCEval$ (64\%)) beat the best performing handcrafted metric METEOR, which has 61.1\% accuracy in this category. $LCEval$ in particular achieves the best accuracy , which highlights the fact that using all features is helpful in creating metrics which can differentiate between two good quality correct captions.

Table~\ref{table:abstract505} shows that amongst the learned measures, the variant of $LCEval$, which is based on syntactic features, achieves the best performance in HHC category. This also validates the effectiveness of the syntactic features in differentiating between the two good quality captions.  $LCEval$ does not only show the best performance in HHI category, but also achieves the second best average accuracy of 76.13\%. We also observe that \textit{Semantic} achieves a strong performance in HHI category, which can be intuitively explained by the fact that in order to differentiate between a correct and an incorrect caption, semantics play an important role. This is one of the reasons why metrics that are based on semantic features show a strong performance in HHI category. In the case where both captions are correct, resorting to the similarity between the syntactic structure of the reference and candidates can be useful. This is why we observe that syntactic features prove to be the most useful in this case.  

\begin{table*}[t]
	\footnotesize
	\centering
	\caption{Comparative accuracy results on various distraction tasks}
	\vspace{-1em}
	\label{table:hodosh}
	\begin{tabular}{|m{1.8cm}|*{7}{c|}}
		\hline
		\textbf{Metrics}   & Replace Person & Replace Scene & Share Person  & Share Scene   & Just Person   & Just Scene    & AVG          \\ \hline
		$\text{BLEU}_1$            & 84.9           & 78.1          & 87.5          & 88.2          & 87.5          & 98.4          & 87.4                        \\
		\rowcolor{Gray}
		$\text{BLEU}_4$            & 85.9           & 75.2          & 83.5          & 82.4          & 54.9          & 67.7          & 72.1                       \\
		$\text{ROUGE}_L$           & 83.3           & 71.1          & 86.8          & 86.8          & 83.4          & 94.1          & 84.1                        \\
		\rowcolor{Gray}
		METEOR            & 83.7           & 75.1          & 92.4          & 91.4          & 91.9          & 98.4          & 89.3                       \\
		$\text{CIDEr}_D$             & 89.9  & \textbf{95.0} & 94.1          & 93.1          & 73.3          & 81.5          & 85.7                        \\
		\rowcolor{Gray}
		SPICE             & 84.0           & 76.0          & 88.5          & 88.8          & 78.1          & 92.0          & 83.6                      \\
		SPIDEr            & 89.7           & \textbf{95.0} & 94.7          & 93.6 & 76.6          & 86.1          & 89.3    \\
		\rowcolor{Gray}                   
		WMD            & 87.6           & 87.8		 & 94.3          & 92.5 & 86.3          & 96.9          & 90.9   \\        
		\hline \hline
		\rowcolor{Gray}
		Semantic          & 86.3           & 78.5          & 95.1          & 92.9          & 84.8          & 97.3          & 89.2                        \\
		Syntactic         & 61.2           & 48.2          & 87.1          & 84.8          & 79.1          & 88.8          & 74.9                        \\
		\rowcolor{Gray}
		Lexical           & 60.1           & 80.1          & 88.2          & 86.3          & \textbf{94.8} & \textbf{98.7} & 84.7                        \\
		Sem-Syn           & 89.2           & 88.5          & \textbf{95.4} & 93.3          & 83.2          & 93.0          & 90.4                       \\
		\rowcolor{Gray}
		Sem-Lex           & 80.7           & 85.6          & 93.0          & 91.1          & 94.3          & 98.6          & 90.7                        \\
		Syn-Lex           & 76.0           & 82.4          & 90.9          & 89.3          & 90.4          & 94.8          & 87.3                        \\
		\rowcolor{Gray}
		NNEval            & \textbf{90.2}           & 91.8          & 95.1          &\textbf{94.0}          & 85.8          & 94.7          & \textbf{91.9}                        \\
		\textbf{$LCEval$} & 81.2           & 85.9          & 92.7          & 91.4          & 92.1          & 97.8          & 90.2                        \\ \hline
		\#Instances       & 5,816           & 2,513          & 4,594          & 2,619          & 5,811          & 2,624          & Total: 23,977 \\ \hline
	\end{tabular}
	%\end{center}
\end{table*}

As evident from the results in Table~\ref{table:pascal} and Table~\ref{table:abstract505}, $LCEval$ outperforms the handcrafted metrics on average, taking the lead amongst all the metrics in the most difficult category i.e., HHC of PASCAL50S. This shows that a careful selection of features can lead to a better customised learned metric.

\subsection{Robustness Analysis}

To evaluate the robustness of the learned metrics against various sentence perturbations, we use HODOSH dataset (see Sec. 4.5.3) which consists of images, each paired with two candidate captions, one correct and the other incorrect (distracted version of correct caption). A robust image captioning metric should choose the correct over the distracted one.

The average accuracy scores for the evaluation metrics are reported in Table~\ref{table:hodosh}. The last row of Table~\ref{table:hodosh} shows the numbers of instances tested for each category. Overall, it can be observed that the learned metrics outperform the handcrafted measures in terms of average accuracy (NNEval 91.9\%). Whereas, amongst the handcrafted metrics, WMD achieves the highest average accuracy.

\begin{figure*}[t]
	\centering
	\includegraphics[width=1\textwidth]{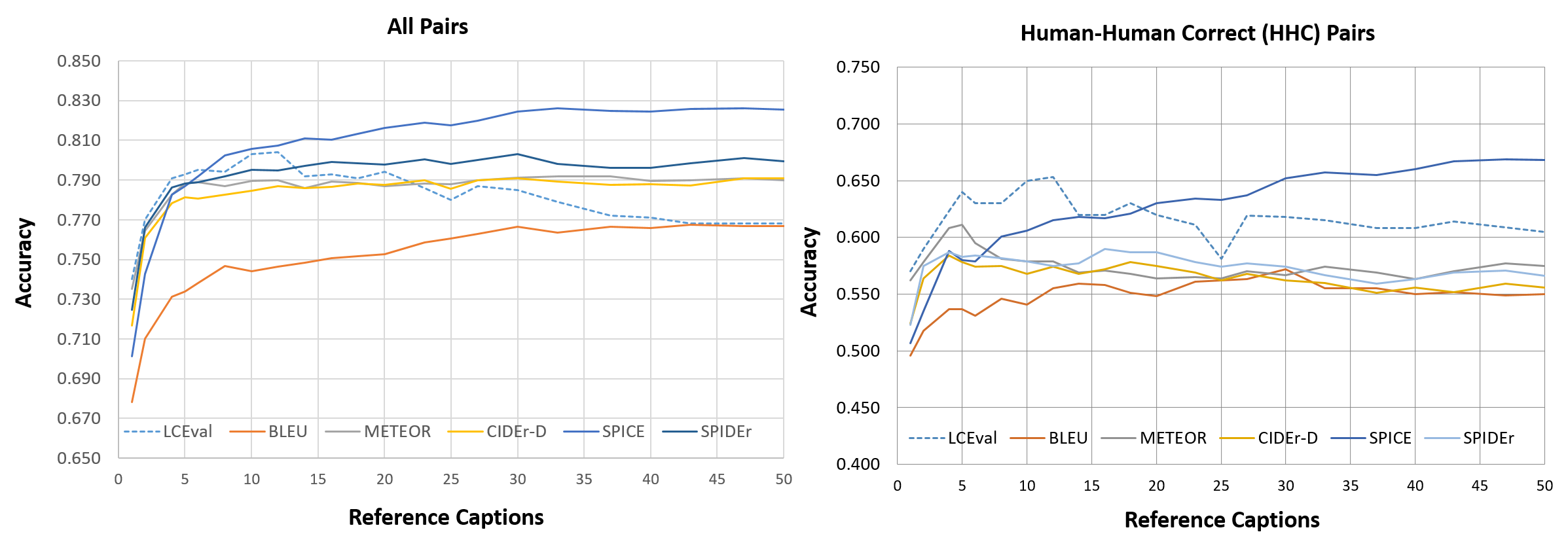}
	\vspace{-1.5em}
	\caption{Accuracy graphs with a variable number of reference captions for different metrics.}
	\label{fig:PascalVariation}
	%\vspace{-0.5em}
	%\end{center}
\end{figure*}
NNEval achieves the highest accuracy when the perturbation is only limited to the mention of a person, (i.e., Just-Person category) . Whereas, $\text{CIDEr}_D$ and SPIDEr achieve the best performance in Just-Scene category. In Share Person/Scene category, where every word in the caption except for the mentioned Person/Scene is different (or wrong), the change is easily picked up by all the measures. Learned \textit{Semantic} measures are the most robust to such a perturbation, compared to \textit{Lexical} and \textit{Syntactic} learned measures. Though combining semantic and syntactic measures (\textit{Sem-Syn})) further adds to their robustness achieving 95.4\% and 93.3\% in the case of Share-Person and Share-Scene categories, respectively. In both Just-Person and Just-Scene categories, the learned \textit{Lexical} metric outperforms all the others, because the perturbed captions are based on noun phrase, and the n-gram precision-based features prefer captions which have a higher n-gram overlap with the reference. 

Note that NNEval is the most consistent performer among all metrics. Overall, $LCEval$ also shows a decent performance, and is more robust compared to NNEval in the Just-Scene and Just-Person categories. The results in Table~\ref{table:hodosh} show that \textit{Sem-Lex} achieves an overall accuracy of 90.7, which is lower compared to the accuracy of NNEval i.e., 91.9. As mentioned before, NNEval can be regarded as a special case of the \textit{Sem-Lex} category. The main differences in features between \textit{Sem-Lex} and NNEval are MOWE and unigram recall, which are both Bag-of-Word (BOW) based measures. Such measures are not that robust to changes in the word order. As long as the candidate sentence contains the same words as the reference caption, it is given a higher score by these measures (unigram precision and recall, MOWE, etc.,), regardless of the word order. On the other hand, we observe that for Just-Person and Just-Scene categories, the performance of unigram BOW like methods is high, for example $\text{BLEU}_1$, WMD and METEOR. This is because, word-matching based measures such as $\text{BLEU}_1$ penalize captions that do not contain the lexicons present in the reference captions. Addition of unigram measures can be one of the reasons for the lower robustness of $LCEval$ compared to NNEval. Thus, we emphasize the importance of careful feature engineering in order to make a strong learned composite measure. Moreover, the shortcomings (of learned metrics), can be rectified by adding perturbed negative training examples in the training set, to help the metrics pay more attention to these fine details.

\subsection{Impact of the Number of Reference Captions}

In this experiment, we investigate the impact of the number of reference captions on the accuracy of the proposed learned metric. For this purpose, we use PASCAL50S dataset, which contains 50 reference captions per candidate caption pair. We evaluate the overall accuracy, as well as that of HHC category, since our metric outperforms other metrics in this particular category. We evaluate our proposed metric along with other captioning measures with up to 50 sentences used as a reference. We use a step size of two for this test to reduce the overall overhead of the experiment, in particular, SPICE and METEOR computations are time intensive.

From Figure \ref{fig:PascalVariation}, we observe that the overall accuracy of all the metrics improve for the first 10 sentences. The accuracy of SPICE, SPIDEr and BLEU continue to improve up to 50 references. From 1 to 12 sentences, the accuracy of $LCEval$ improves from 0.740 to 0.804. However, it shows a decline in performance with a higher number of reference captions. Since some of the features that are used for $LCEval$ use a \textit{max} operation over multiple references (i.e., the reference that is the closest to the candidate is selected for the similarity evaluation), the probability of matching a low quality reference increases \cite{vedantam2015cider}. A similar phenomenon can be observed for the HHC accuracy as well in Figure \ref{fig:PascalVariation}, where the accuracy improves from 0.570 to 0.653, by increasing number of references from 1 to 12. However, beyond 12 sentences, the accuracy starts  to decrease. From this experiment, we come to an understanding that our learned metric is sensitive to the quality and number of reference captions used. Moreover, using the \textit{max} operation over the reference captions limits the ability of the metrics to exploit a larger number of reference captions. In future work, we would like to investigate the impact of using \textit{max}, \textit{min} and \textit{mean} across references (for the features of our metric) in detail. See Appendix A.2 for our preliminary results on using \textit{min} and \textit{mean} operation over the reference captions.

\begin{table*}[h]
	\centering
	\caption{Kendall's {$\tau$} correlation of metrics trained using different machine examples}
	\vspace{-1.0em}
	\label{table:training Variation}
	\begin{tabular}{|m{2.5cm}||M{4cm}|M{3.5cm}|M{2.5cm}|M{2cm}|}
		\hline
		\textbf{Metric}                  & \textbf{Machine Model}     & \textbf{Source}        & \textbf{Total Training Examples} & \textbf{Kendall Correlation} \\ \hline
		$LCEval$ (AA)                    & Adaptive Attention (AA)    & Lu et al., 2017 \cite{lu2017knowing}   & 88,656                    & 0.403                \\
		\rowcolor{Gray}
		$LCEval$ (SAT) & Show Attend and Tell (SAT) & Xu et al., 2015 \cite{xu2015show}      & 88,656                    & 0.417                \\
		$LCEval$ (ST)                    & Show and Tell (ST)         & Vinyals et al., 2015 \cite{vinyals2015show} & 88,656                    & \textbf{0.421}       \\
		\rowcolor{Gray}
		$LCEval$       & AA+ST+SAT                  & \cite{lu2017knowing}, \cite{vinyals2015show}, \cite{xu2015show}                     & 132,984                   & 0.418                \\ \hline
	\end{tabular}
	%\end{center}
\end{table*}

\begin{table}[h]
	\centering
	\caption{Comparative accuracy results on four kinds of caption pairs tested on PASCAL50S}
	\vspace{-1.0em}
	\label{table:pascalTrain}
	\begin{tabular}{|m{2.5cm}|*{4}{c|}}
		\hline
		\textbf{Metric}            & \textbf{HHC}   & \textbf{HM}   & \textbf{MM}   & \textbf{AVG}  \\ \hline
		$LCEval$ (AA)                    & 60.1          & 89.7          & 67.7          & 78.9         \\
		\rowcolor{Gray}
		$LCEval$ (SAT)   & 58.7          & 85.2          & 69.7          & 77.8          \\
		$LCEval$ (ST)                    & 61.6          & 86.7          &\textbf{ 69.7}          & 79.0          \\
		
		\rowcolor{Gray}
		\textbf{$LCEval$}          & \textbf{64.0} & \textbf{90.1}          & 65.4          & \textbf{79.3}          \\ \hline
	\end{tabular}
	%\end{center}
\end{table}

\subsection{Impact of Training Examples}

In this experiment, we compare the performance of metrics that are developed using different training examples, machine examples in particular. As described in Section 4.3, we use the captions of three machine approaches in our training set. The captions generated by these approaches vary in quality, where ``Adaptive Attention'' (AA) captioning system \cite{lu2017knowing} has the best reported performance and ``Show and Tell'' (ST) \cite{vinyals2015show} attains the lowest evaluation score amongst the three. We create three separate training sets, each containing the machine captions of only one of the three approaches. We keep the human captions of these smaller training sets the same as the ones in our larger set, so that we can study the influence of machine examples on the learned metric. We evaluate the caption-level correlation of the metrics on the COMPOSITE set (Section 4.5.1) and report the Kendall's {$\tau$} correlation coefficient in Table~\ref{table:training Variation}. 

Although the metrics shown in Table~\ref{table:training Variation}, $LCEval$ (AA), $LCEval$ (SAT), and $LCEval$ (ST),  were trained using only 66\% of the data which was used to train $LCEval$, the smaller training set size did not significantly impact on the caption-level correlation of the trained metric. Therefore, we can say that the quality of the training examples is more important than quantity. Moreover, the results suggest that it is helpful to include sentences from bad systems. That is, having seen only good examples is more harmful than not having seen training examples from a bad system.

As far as accuracy is concerned (See Table~\ref{table:pascalTrain}), training on only good examples decreases the ability of a metric to distinguish between machine-generated captions. Whereas, training on a good variety of machine generated captions improves the HM and HHC classification accuracy. 

Moreover, increasing the variety of examples would give $LCEval$ the opportunity to learn better. To ensure that the $LCEval$ remains an effective predictor for modern systems as they continue to improve, new training examples need to be added to the training set, potentially with more powerful linguistic features for the metric to be re-trained. Though, this would imply that as the metric is retrained over time, its scores will hold little absolute meaning, but the advantages in regards to the fine-grained error analysis and adaptability might be substantial.

\subsection{Timing Analysis}

In this experiment we analyze the execution time of the proposed metric $LCEval$, against the various existing metrics. The running times where recorded on 4000 captions of PASCAL50S test set. The evaluation scores of all the metrics were evaluated using one reference per candidate caption. We performed this experiment on Core-i7, 3.5 GHz CPU with 64GB DDR3 RAM. We used the MSCOCO evaluation scripts for metrics such as BLEU, ROUGE, CIDEr, METEOR and SPICE. Whereas, for WMD we used the Gensim library \cite{khosrovian2008gensim}. The only metrics that we implemented are n-gram precision, unigram recall, MOWE and HWCM. Amongst the handcrafted measures, simple metrics such as ROUGE and BLEU are the least time consuming, whereas complex metrics such as HWCM and SPICE, which require linguistic parsers are the most time expensive.

The two learning bases measures that we analysed are $NNEval$ and $LCEval$. We split the execution times of these metrics into two phases: 1) feature extraction and 2) score generation. $LCEval$-feature extraction represents the cumulative time that is required to extract the scores of various measures that are used by the $LCEval$ model as features. Whereas, $LCEval$-score generation indicates the time that is required to generate the metric scores using extracted features. We can observe that the feature extraction time for $LCEval$ is 1.8 times more than that of $NNEval$. This is due to the fact that the number of features used by $LCEval$ are 3 times more than that of $NNEval$. However, the score computation for $LCEval$ is about 1.7 times faster than $NNEval$. This is mainly due to the fact that the network used for $LCEval$ is much simpler (12 nodes, 1 hidden layer), compared to $NNEval$ (72-72 nodes, 2-hidden layers).

We do realize that handcrafted metrics are less time expensive than the learned ones, but that comes at the cost of lower caption and system-level correlation (see Sec. 5.2 and 5.8). The main bottleneck in learned measures is the feature extraction phase, which can be optimized either by using less time expensive features or using more optimized implementations. However, the focus of this work was on maximizing the correlation rather than optimizing the feature extraction process for lower computational costs.
 
\begin{table}[t]
	\centering
	\caption{Execution times of various metrics on PASCAL50S dataset}
	\vspace{-1.0em}
	\label{table:NewTab1}
	\begin{tabular}{|m{4cm}||M{2cm}| }
		\hline
		\textbf{Metric}                              & \textbf{Time (secs)} \\ \hline
		METEOR                                       & 007.92                 \\
		\rowcolor{Gray}
		SPICE                      & 095.60                \\
		$\text{BLEU}_4$                              & 000.32                 \\
		\rowcolor{Gray}
		N-gram Precision ($0<n<5$) & 000.30                 \\
		Unigram Recall                               & 000.10                 \\
		\rowcolor{Gray}
		$\text{ROUGE}_L$           & 000.16                 \\
		$\text{CIDEr}_D$                             & 015.73                \\
		\rowcolor{Gray}
		WMD                        & 003.23                \\
		MOWE                                         & 045.50                \\
		\rowcolor{Gray}
		HWCM                       & 060.30                \\
		NNEval-score generation                               & 000.63                 \\
		\rowcolor{Gray}
		NNEval-feature extraction  & 122.73               \\
		$LCEval$-score generation                               & 000.37                 \\
		\rowcolor{Gray}
		$LCEval$-feature extraction  & 228.83               \\ \hline
	\end{tabular}
	%\end{center}
\end{table}
\begin{table*}[h]
	\centering
	\caption{Pearson's $p$ correlation of various evaluation metrics against 12 competition entries on the COCO validation set \cite{cui2018learning}. Our reported scores with a * differ from the ones reported in \cite{cui2018learning} M1: Percentage of captions that are evaluated as better or equal to human caption.
		M2:	Percentage of captions that pass the Turing Test.
		M3:	Average correctness of the captions on a scale of 1-5 (incorrect - correct).
		M4:	Average amount of detail of the captions on a scale of 1-5 (lack of details - very detailed).
		M5:	Percentage of captions that are similar to human description.
	}
	\vspace{-1.0em}
	\label{table:system level}
	%\scriptsize
	\begin{tabular}{|m{1.7cm}|M{1.1cm}M{1cm}|M{1.1cm}M{1.1cm}|M{1.1cm}M{1.1cm}|M{1.1cm}M{1.1cm}|M{1.1cm}M{1.1cm}|}
		\hline
		 & \multicolumn{2}{c|}{\textbf{M1}} & \multicolumn{2}{c|}{\textbf{M2}} &\multicolumn{2}{c|}{\textbf{M3}} &\multicolumn{2}{c|}{\textbf{M4}} &\multicolumn{2}{c|}{\textbf{M5}}  \\
		
		\textbf{Metric} & $p$ & p-value &   $p$ &  p-value &   $p$  & p-value &  $p$ &  p-value &    $p$ &  p-value\\ \hline
		
		$\text{BLEU}_1$	&	0.12 &		0.69 &	0.14 &	0.66 &	0.55 &	0.05 &	-0.52 &	0.07 &	0.24 &	0.43 \\
		\rowcolor{Gray}
		$\text{BLEU}_4$	&	-0.02 &	0.95 &	-0.01 &	0.99 &	0.46 &	0.11 &	-0.58 &	0.04 &	0.14 &	0.65 \\
		$\text{ROUGE}_L$ &	0.09 &		0.77 &	0.10 &	0.75 &	0.53 &	0.06 &	-0.53 &	0.07 &	0.21 &	0.49 \\
		\rowcolor{Gray}
		METEOR	&	0.61 &		0.03 &	0.59 &	0.03 &	0.81 &	0.00 &	0.09 &	0.78 &	0.69 &	0.01  \\
		$\text{CIDEr}_D$ &	0.47* &		0.13* &	0.50* &	0.10* &	0.86* &	0.00* &	-0.15* &	0.63* &	0.53* &	0.08* \\
		\rowcolor{Gray}
		SPICE	&	0.76* &		0.00* &	0.75* &	0.01* &	0.79* &	0.00* &	0.16* &	0.42* &	0.74* &	0.01* \\
		WMD	&	0.68 &		0.02 &	0.71 &	0.01 &	\textbf{0.89} &	0.00 &	-0.28 &	0.37 &	0.72 &	0.01 \\
		\rowcolor{Gray}
		LIC/LIC(DA)	&	0.82/0.94 &	0.00 &	0.81/0.96 &	0.00 &	0.43/0.72 &	0.14/0.01 &	\textbf{0.84}/0.63 &	0.00/0.03 &	0.70/0.87 &	0.01/0.00 \\
		NNEval	&	0.69 &		0.01 &	0.69 &	0.01 &	0.84 &	0.00 &	-0.22 &	0.49 &	0.69 &	0.01 \\
		\rowcolor{Gray}
		$LCEval$	&	\textbf{0.86} &		0.00 &	\textbf{0.82} &	0.00 &	0.70 &	0.01 &	0.18 &	0.59 &	\textbf{0.78} &	0.00 \\
		\hline
	\end{tabular}
	%\end{center}
\end{table*}

\subsection{System-Level Correlation}

We also evaluate the system-level correlation of our metric to compare various handcrafted and learned metrics. For this experiment, we use human judgements collected in the 2015 COCO Captioning Challenge for 12 teams who participated in this captioning challenge. We report M1: Percentage of captions that are evaluated as better or equal to human caption, M2: Percentage of captions that pass the Turing Test, M3: Average correctness of the captions on a scale of 1-5 (incorrect - correct), M4: Average amount of detail of the captions on a scale of 1-5 (lack of details - very detailed) and M5: Percentage of captions that are similar to human description. While M1 and M2 were used to rank the captioning models in the COCO challenge, M3, M4 and M5 were intended for an ablation study to understand the various aspects of caption quality.

The human judgements were collected on MSCOCO test set, which is not publicly available. Therefore, as done in the literature \cite{cui2018learning} we perform our experiments on the COCO validation set. We use the captions of 12 captioning systems, who submitted their results on the validation set for the 2015 COCO captioning challenge. We also assume that the human evaluations on test set and validation set are fairly similar, as assumed in \cite{cui2018learning}.

In the Table~\ref{table:system level}, we report Pearson's $p$ to keep the choice of correlation coefficient consistent with \cite{anderson2016spice} and \cite{cui2018learning} for scoring captioning metrics. Kendall and Spearman rank correlations reflect the similarity of the pairwise rankings whereas Pearson's $p$ captures the linear association between data points. Following the guidance of Dancey and Reidy \cite{dancey2004statistics} to classify the strength of correlations, which states that a coefficient of 0.0-0.1 is uncorrelated, 0.11-0.4 is weak, 0.41-0.7 is moderate, 0.71-0.90 is strong,  and 0.91-1.0 is perfect, $LCEval$ shows a strong correlation with human judgments in both M1 and M2 categories. We also report the correlation of a recently proposed learned measure LIC \cite{cui2018learning} , which shows a competitive performance. However, LIC(DA) was trained on one half of the MSCOCO validation set using data augmentation, and evaluated on the other half. $LCEval$ is trained on Flickr30k dataset without any data augmentation and performs better than all other measure as well as LIC (trained without data augmentation),  in three out of five categories.

\section{Conclusion and Future Work}

We propose in this paper $LCEval$, a learned metric which measures the quality of a caption based on various linguistic aspects. Our empirical results demonstrate that $LCEval$ correlates with human judgements better than the existing metrics at the caption level, and also achieves a strong system-level correlation. Our proposed framework, facilitated the incorporation of various useful features which contributed towards the successful performance of our metric. We carried out a detailed analysis of the impact of various features on the caption-level correlation, the classification accuracy and robustness. Our results show that with careful feature engineering, task specific learned metrics can be successfully designed.

We tested three types of linguistic features and out of these, the lexical features were found to be the most influential in the improvement of the caption-level correlation. The semantic and the syntactic features also carry useful information, and using only these for the proposed learned metric yields a correlation of 0.366 on the test set, which is higher than the correlation of individual handcrafted features. Used together, all three features lead to a much stronger metric, compared to the handcrafted measures. However, our ablation analysis suggests that there might be some informational redundancies in the full feature set, which can be investigated in future work. 

In terms of differentiating a pair of captions on the basis of quality, no single metric, (either learned or handcrafted) was able to perform consistently well in all of the tested categories. This shows that features have to be engineered for each specific task if a high performance is required for that particular task. The same holds true for robustness, as no learned metric consistently outperforms all the others, in all of the categories. This highlights a limitation of the learned metrics i.e., they are not robust to all different types of sentence perturbations. These shortcomings of the learned metrics can be resolved by adding perturbed negative training examples in the dataset, to help the metrics pay more attention to the fine details. Training examples  play a significant role in the improvement of the robustness to sentence perturbations, and towards the overall performance of $LCEval$. Our analysis also revealed that the quality of the training examples is more important than its quantity. 

An important aspect of learning approach to caption evaluation is customization, which learned measures can provide on top of improving the correlation with human judgements. Features can be added or fine-tuned to specifically improve the metrics sensitivity towards certain aspects of caption quality such as adequacy, fluency or acceptability. Moreover, as the captioning systems continue to improve, more powerful features can be incorporated or designed to maintain or even improve the sensitivity of the metric to distinguish between captions. On top of that, adding training examples from newer systems will also improve the discriminatory power of the metric.

In future, we plan to investigate the performance of our proposed metric on multilingual captions. We have released our code\footnote{https://github.com/NaehaSharif/LCEVal} and hope that it will lead to further development of learning-based evaluation metrics and contribute towards fine-grained assessment of captioning models.

\section*{Acknowledgements}
We are grateful to NVIDIA for providing Titan-Xp GPU, which was used for the experiments. We would like to thank Somak Aditya for sharing COMPOSITE dataset and Ramakrishna Vedantam for sharing PASCAL50S and ABSTRACT50S datasets. We would also like to thank Yin Cui for providing us the dataset containing the captions of 12 teams that participated in the 2015 COCO captioning challenge.  \\
This work is supported by Australian Research Council, ARC DP150100294.
\clearpage
         
\bibliographystyle{splncs04}
\bibliography{mybib}
\bigskip

\clearpage
\appendix
\section{Appendix}

\subsection{Performance on Machine Translations}

To analyse the performance of our learned metric on sentences from a different domain, we experiment with the Machine Translation (MT) dataset. The MT dataset used in WMT shared metrics task \cite{ma2018results} are publicly available along with the human judgements. The metrics are required to evaluate the translation quality based on the linguistic comparison against reference translations. To the best of our knowledge, such analysis has never been performed. For this experiment we use a dataset from WMT18 Metrics Shared Task \cite{ma2018results}  i.e., Direct Assessment (DA) segment-level newstest2016 which contains segment-level human scores for Machine Translations.

We focus on tasks involving translation to English from other languages. DAseg WMT16 contains translations from 6 source languages: Finnish (FI), Czech (CS), Russian (RU), German (DE), Roman (RO), Turkish (TR), containing 560 sentences each. We assess our metrics in terms of Kendall's correlation with human judgements across different languages. We also evaluate the macro average of Kendall correlation across different languages. The human quality judgements are based on the similarity between the reference and candidate translation, the more similar it is to the reference the better it is in quality.  

It can be seen that machine translation metric METEOR achieves the highest sentence-level correlation with human judgements across all the languages. Whereas, captioning specific metrics show a comparatively lower performance. Amongst the captioning specific metrics $LCEval$, or its variants show the best sentence-level correlation on most of the language pairs. One particular reason for this is that $LCEval$ is a learned-composite measure and we believe that its performance could be improved further by training it on Machine Translations.

\begin{table*}[t]
	\centering
	\caption{Sentence-level Kendall correlation on WMT shared metrics segment-level newstest2016. All p-values are less than 0.001}
	\vspace{-1.0em}
	\label{table:machine}
	\begin{tabular}{|m{1.5cm}|*{7}{c|}}
		\hline
		\textbf{Metric} & \textbf{CS-EN 
		} &\textbf{DE-EN} &\textbf{FI-EN} &\textbf{RO-EN} &\textbf{RU-EN} &\textbf{TR-EN} &\textbf{Average}\\ \hline
		$\text{BLEU}_1$ &	0.418 &	0.299 &	0.295 &	0.328 &	0.329 &	0.358 &	0.338 \\
		\rowcolor{Gray}
		$\text{BLEU}_4$ &	0.354 &	0.238 &	0.238 &	0.290 &	0.268 &	0.276 &	0.277 \\
		$\text{ROUGE}_L$ &  	0.471 &	0.354 &	0.340 &	0.354 &	0.357 &	0.368 &	0.374 \\
		\rowcolor{Gray}
		METEOR &  	\textbf{0.479} &	\textbf{0.340} &	\textbf{0.378} &	\textbf{0.417} &	\textbf{0.406} &	\textbf{0.424} &	\textbf{0.408} \\
		$\text{CIDEr}_D$ &  	0.398 &	0.266 &	0.293 &	0.330 &	0.316 &	0.360 &	0.327 \\
		\rowcolor{Gray}
		SPICE &  	0.312 &	0.219 &	0.222 &	0.268 &	0.263 &	0.293 &	0.263 \\
		WMD &  	0.375 &	0.270 &	0.286 &	0.312 &	0.310 &	0.324 &	0.313 \\
		\rowcolor{Gray}
		SPIDER & 	0.403 &	0.283 &	0.287 &	0.338 &	0.338 &	0.363 &	0.336 \\
		\hline
		Semantic &	0.371 &	0.272 &	0.281 &	0.31 &	0.327 &	0.353 &	0.320 \\
		\rowcolor{Gray}
		Syntactic &	0.411 &	0.315 &	0.33  &	0.37 &	0.309 &	0.315 &	0.340 \\
		Lexical &	0.437 &	0.288 &	0.339 &	0.383 &	0.368 &	0.385 &	0.370 \\
		\rowcolor{Gray}
		Sem-Syn &	0.356 &	0.258 &	0.265 &	0.294 &	0.312 &	0.341 &	0.300 \\
		Sem-Lex &	0.412 &	0.285 &	0.311 &	0.361 &	0.363 &	0.374 &	0.350 \\
		\rowcolor{Gray}
		Syn-Lex &	0.432 &	0.288 &	0.339 &	0.395 &	0.361 &	0.368 &	0.360 \\
		$LCEval$ &	0.419 &	0.294 &	0.318 &	0.364 &	0.363 &	0.377 &	0.356 \\
		\hline
	\end{tabular}
	%\end{center}
\end{table*}

\begin{figure}[t]
	\centering
	\includegraphics[width=0.5\textwidth]{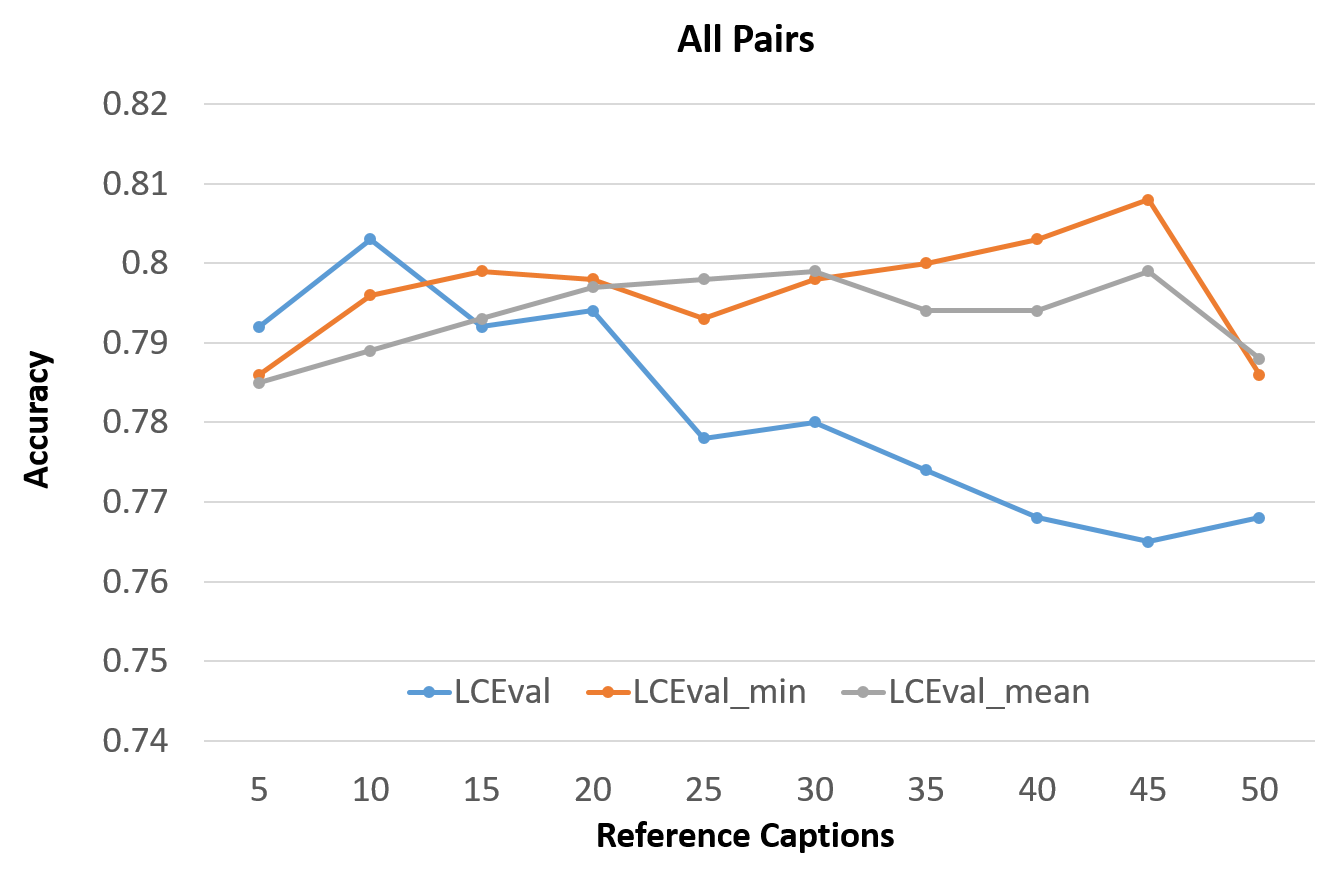}
	\vspace{-1.5em}
	\caption{Accuracy graphs with a variable number of reference captions for $LCEval$, $\text{$LCEval$}_{min}$ and $\text{$LCEval$}_{mean}$. $LCEval$, uses max operation over the reference captions, whereas $\text{$LCEval$}_{min}$ and $\text{$LCEval$}_{mean}$ use min and mean operation respectively. It is evident that using a 'min' or 'mean' operation helps the metric exploit a larger number of reference captions compared to the 'max' operation.}
	\label{fig:CorVsAcculc}
	%\vspace{-1em}
	%\end{center}
\end{figure}
\subsection{Impact of using \textit{min} and \textit{mean} operation over the reference captions}

We come to an understanding that $LCEval$ is sensitive to the quality and number of reference captions used. Moreover, some of the metrics whose scores are used as features in $LCEval$, use `max' operation over the reference captions, which limits their ability to exploit a larger number of reference captions. Therefore, as a proof of concept, we trained two different versions of $LCEval$: $\text{LCEval}_{min}$ and $\text{LCEval}_{mean}$. For $\text{LCEval}_{min}$ we used `min' operation over the reference captions for all the following features: n-gram precision, unigram recall, HWCM and MOWE. We do not change the implementations of other metrics and leave them as they are. For $\text{LCEval}_{mean}$ we use `mean' operation over the reference captions. Whereas, $LCEval$ uses the `max' operation over the reference.

 We test the trained metrics on the PASCAL50S dataset by varying the number of references. For this experiment we keep the step size for the reference sentences equal to five. Figure \ref{fig:CorVsAcculc} shows the results of our experiment. It is evident that using a `min' or `mean' operation helps the metric exploit a larger number of reference captions compared to the `max' operation. However, we did not analyse the impact of using a `min' or `mean' operation on correlation and robustness and we leave this for future work.

\begin{table*}[h]
	\centering
	\caption{A comparison of the impact of three different feature groups on the learned metric in terms of  correlation coefficients}
	\vspace{-1.0em}
	\label{table:NewTab3}
	\begin{tabular}{|m{2cm}||*{3}{c}|*{3}{c}|}
		\hline
		& \multicolumn{3}{c|}{\textbf{Remove this group}} & \multicolumn{3}{c|}{\textbf{Use only this group}}  \\ 
		Feature Group &	Pearson &	Spearman &	Kendall &	Pearson &	Spearman &	Kendall \\ \hline
		All features &	- &	- &	- &	0.562 &	0.543 &	0.418 \\
		\rowcolor{Gray}
		Semantic &	0.528 &	0.512 &	0.387 &	0.502 &	0.483 &	0.372 \\
		Lexical &	0.499 &	0.482 &	0.366 &	0.524 &	0.508 &	0.409 \\
		\rowcolor{Gray}
		Syntactic &	0.551 &	0.529 &	0.413 &	0.396 &	0.392 &	0.298 \\
		\hline
	\end{tabular}
	%\end{center}
\end{table*}

\clearpage

\end{document}